\documentclass{article}

\usepackage{amsmath} 			
\usepackage{amsfonts}			
\usepackage{amsthm} 			
\usepackage{graphicx} 		
\usepackage{algorithm}
\usepackage{algorithmic} 	
\usepackage{subfigure} 			
\usepackage{url} 					
\usepackage{enumitem}

\usepackage[sort&compress,numbers]{natbib} 

\newcommand{\F}{\mathcal{F}} 
\newcommand{\M}{K,\alpha,T,\eta} 
\newcommand{\G}{\mathcal{G}} 
\newcommand{\E}{\mathbb{E}} 
\newcommand{\evid}{M} 
\newcommand{\mes}{\psi} 
\newcommand{\mess}{\varphi}  
\newcommand{\priortau}{\tau^{(0)}} 
\newcommand{\priormu}{\mu^{(0)}} 

\begin{document}

\title{\large \bf Learning Latent Block Structure in Weighted Networks}


\author{ 
{\sc Christopher Aicher} \\
{\it \small Department of Applied Mathematics, University of Colorado, Boulder, CO, 80309} \\
{\small christopher.aicher@colorado.edu}
\and
{\sc Abigail Z.\ Jacobs} \\
{\it \small Department of Computer Science, University of Colorado, Boulder, CO, 80309}
\and
{\sc Aaron Clauset} \\
{\it \small Department of Computer Science, University of Colorado, Boulder, CO, 80309} \\ 
{\it \small BioFrontiers Institute, University of Colorado, Boulder, CO 80303} \\
{\it \small Santa Fe Institute, Santa Fe, NM 87501}
}
\date{}

\maketitle

\begin{abstract}
{ 
Community detection is an important task in network analysis, in which we aim to learn a network partition that groups together vertices with similar community-level connectivity patterns.
By finding such groups of vertices with similar structural roles, we extract a compact representation of the network's large-scale structure, which can facilitate its scientific interpretation and the prediction of unknown or future interactions.
Popular approaches, including the stochastic block model, assume edges are unweighted, which limits their utility by discarding potentially useful information.
We introduce the \emph{weighted stochastic block model} (WSBM), which generalizes the stochastic block model to networks with edge weights drawn from any exponential family distribution. 
This model learns from both the presence and weight of edges, allowing it to discover structure that would otherwise be hidden when weights are discarded or thresholded. We describe a Bayesian variational algorithm for efficiently approximating this model's posterior distribution over latent block structures. We then evaluate the WSBM's performance on both edge-existence and edge-weight prediction tasks for a set of real-world weighted networks. In all cases, the WSBM performs as well or better than the best alternatives on these tasks.
}
{ 
community detection, weighted relational data, block models, exponential family, variational Bayes.
}
\end{abstract}

\section{Introduction}
Networks are an increasingly important form of structured data consisting of interactions between pairs of individuals in large social and biological data sets. 
Unlike attribute data where each observation is associated with an individual, network data is represented by graphs, where individuals are vertices and interactions are edges. 
Because vertices are pairwise related, network data violates traditional assumptions of attribute data, such as independence.
This intrinsic difference in structure prompts the development of new tools for handling network data. 

In social and biological networks, vertices often play distinct structural roles in generating the network's large-scale structure. 
To identify such latent structural roles, we aim to identify a network partition that groups together vertices with similar group-level connectivity patterns. We call these groups ``communities,'' and their inference produces a compact description of the large-scale structure of a network. (We note that this definition of a ``community'' is more general than the assortative-only definition that is commonly used.) This compact large-scale description itself has many potential uses, including dividing a large heterogeneous system into several smaller and more homogeneous parts that may be studied semi-independently, and in predicting unknown or future patterns of interactions.
By grouping vertices by these roles, community detection in networks is similar to clustering in vector spaces, and many approaches have been proposed~\cite{fortunato:2010}. 

\begin{figure}[tb]
	\centering
		\subfigure{}{
			\includegraphics[width = .22\textwidth, clip = true, trim = 1.5in 1.5in 1.5in 1in]{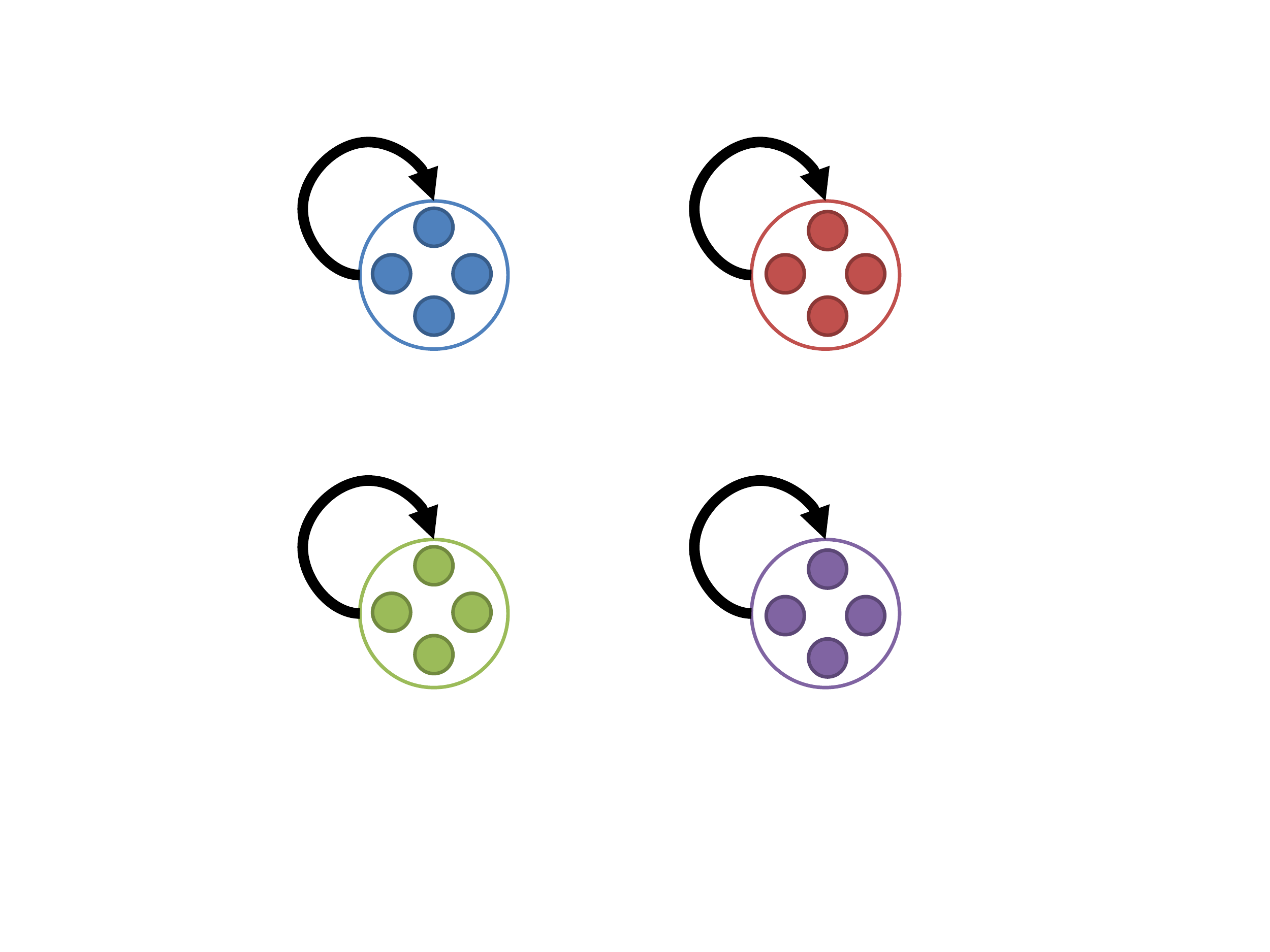}
			\label{fig:AssortativeGraph}
			}
		\subfigure{}{
			\includegraphics[width = .22\textwidth, clip = true, trim = 1.5in 1.5in 1.5in 1in]{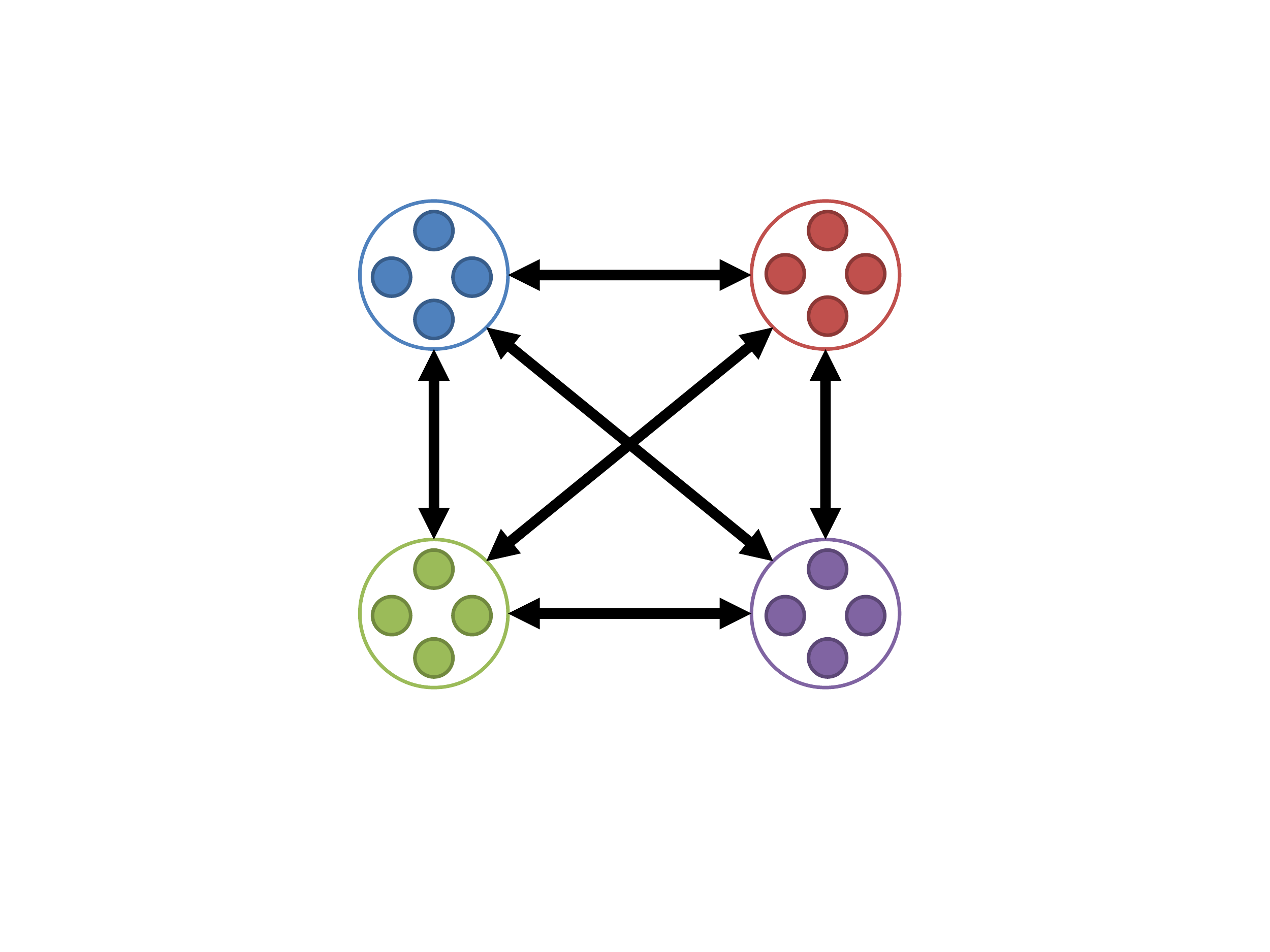}
			\label{fig:DisassortativeGraph}
			}
		\subfigure{}{
			\includegraphics[width = .22\textwidth, clip = true, trim = 1.5in 1.5in 1.5in 1in]{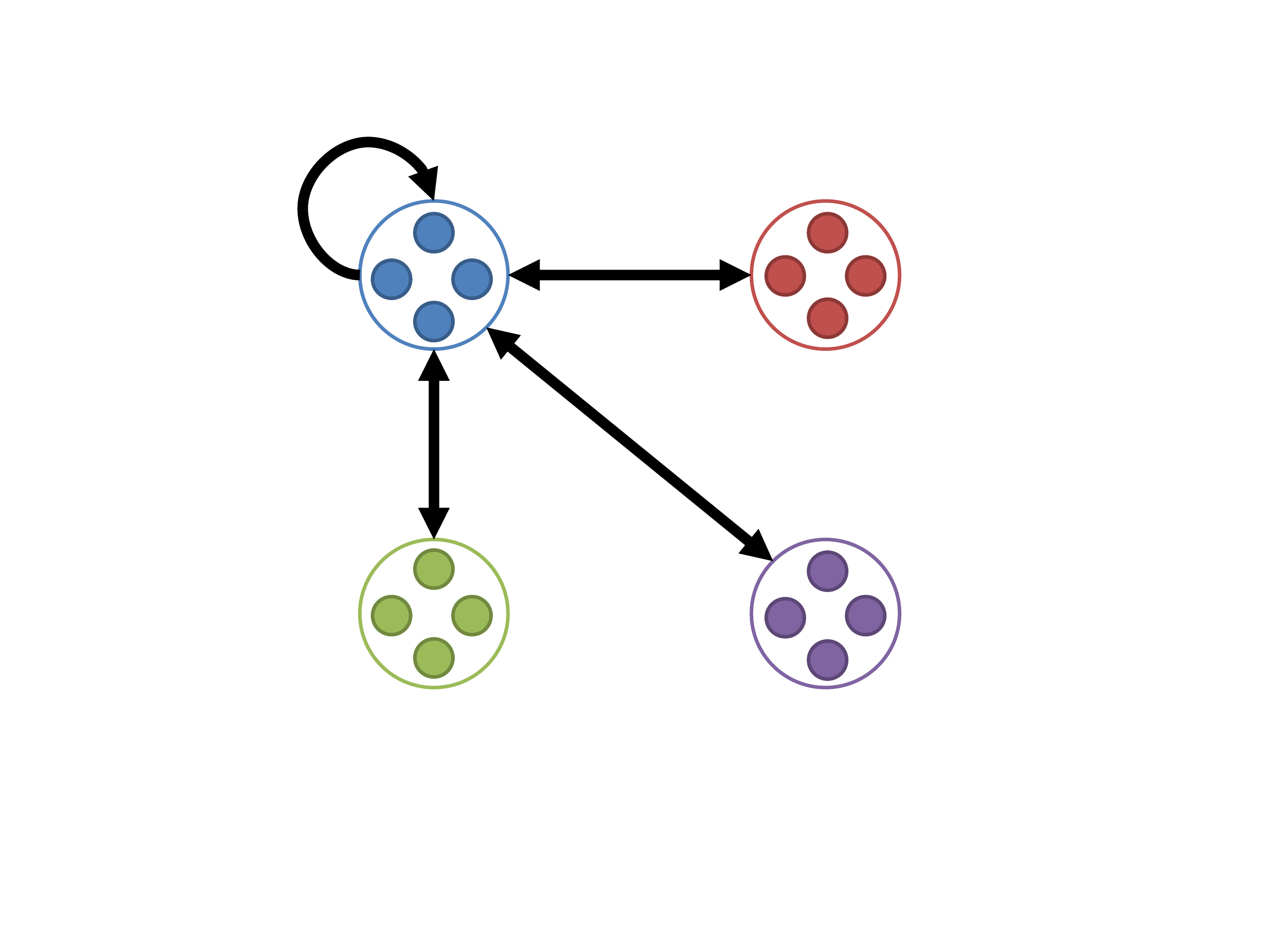}
			\label{fig:CorePeripheryGraph}
			}
		\subfigure{}{
			\includegraphics[width = .22\textwidth, clip = true, trim = 1.5in 1.5in 1.5in 1in]{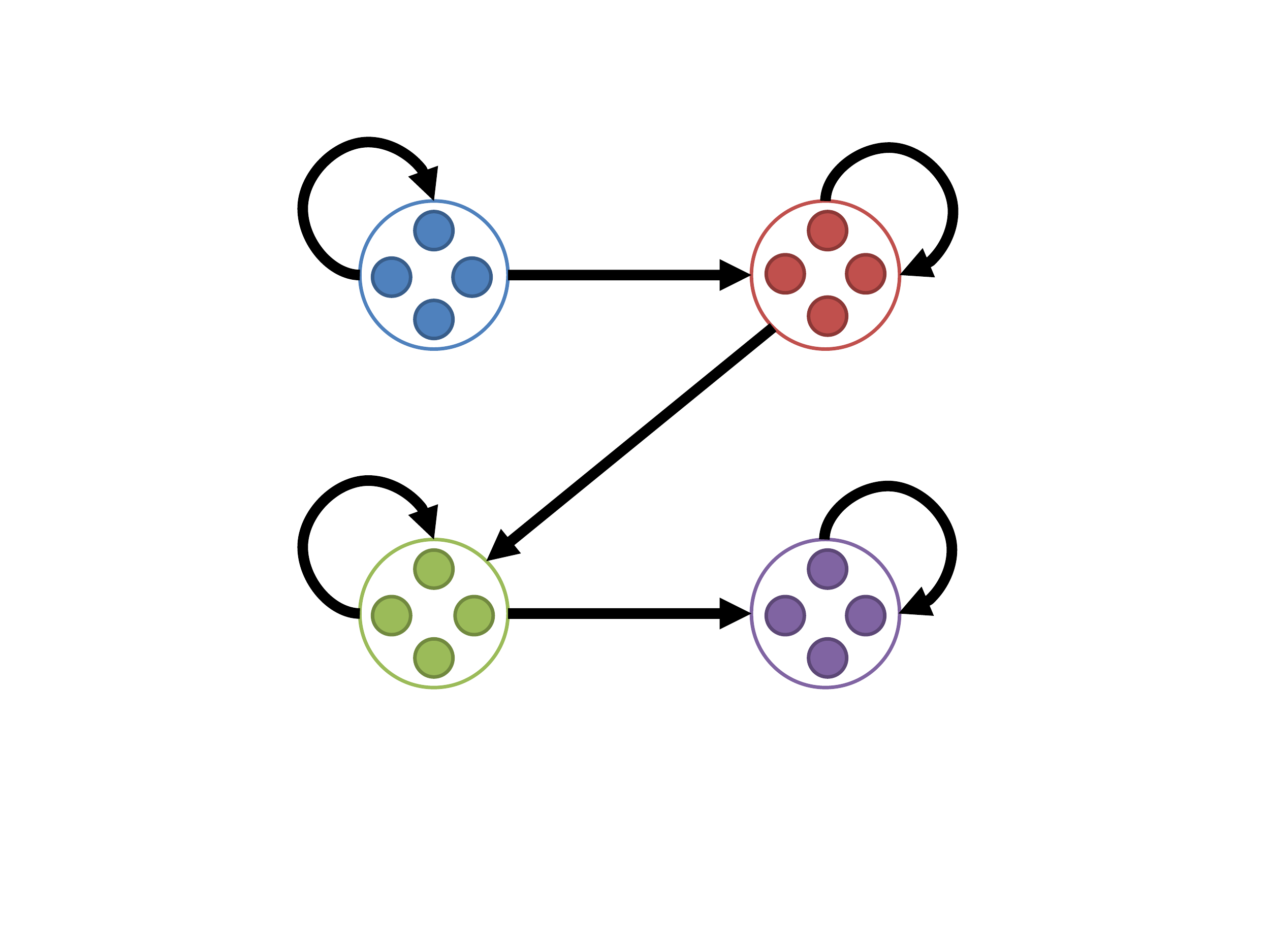}
			\label{fig:OrderedGraph}
			}	
		
		\setcounter{subfigure}{0}
		\subfigure[{Assortative}]{
			\includegraphics[width = .22\textwidth, clip = true, trim = 1.25in 3.1in 1in 3in]{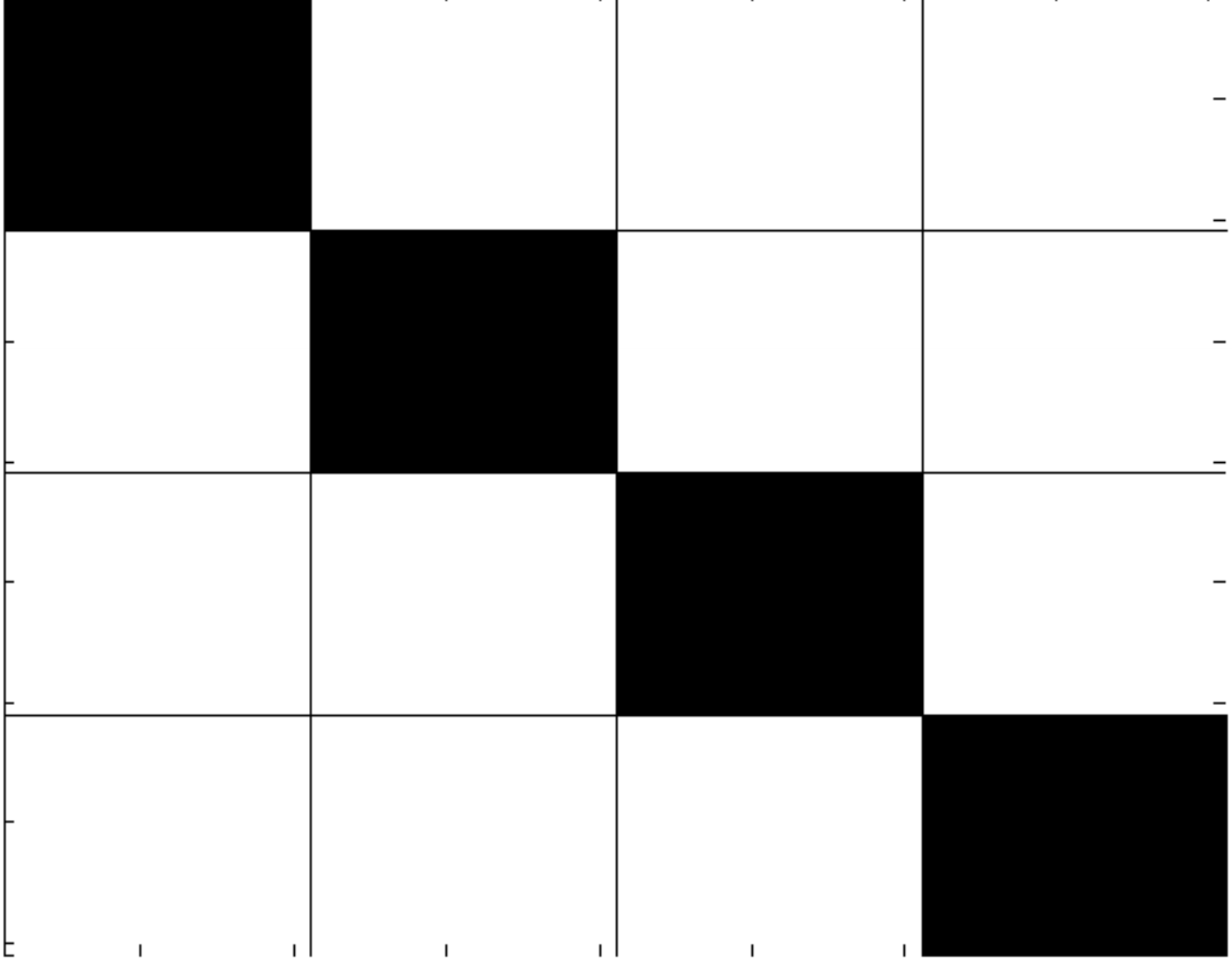}
			\label{fig:AssortativeSBM}
			}
		\subfigure[{Disassortative}]{
			\includegraphics[width = .22\textwidth, clip = true, trim = 1.25in 3.1in 1in 3in]{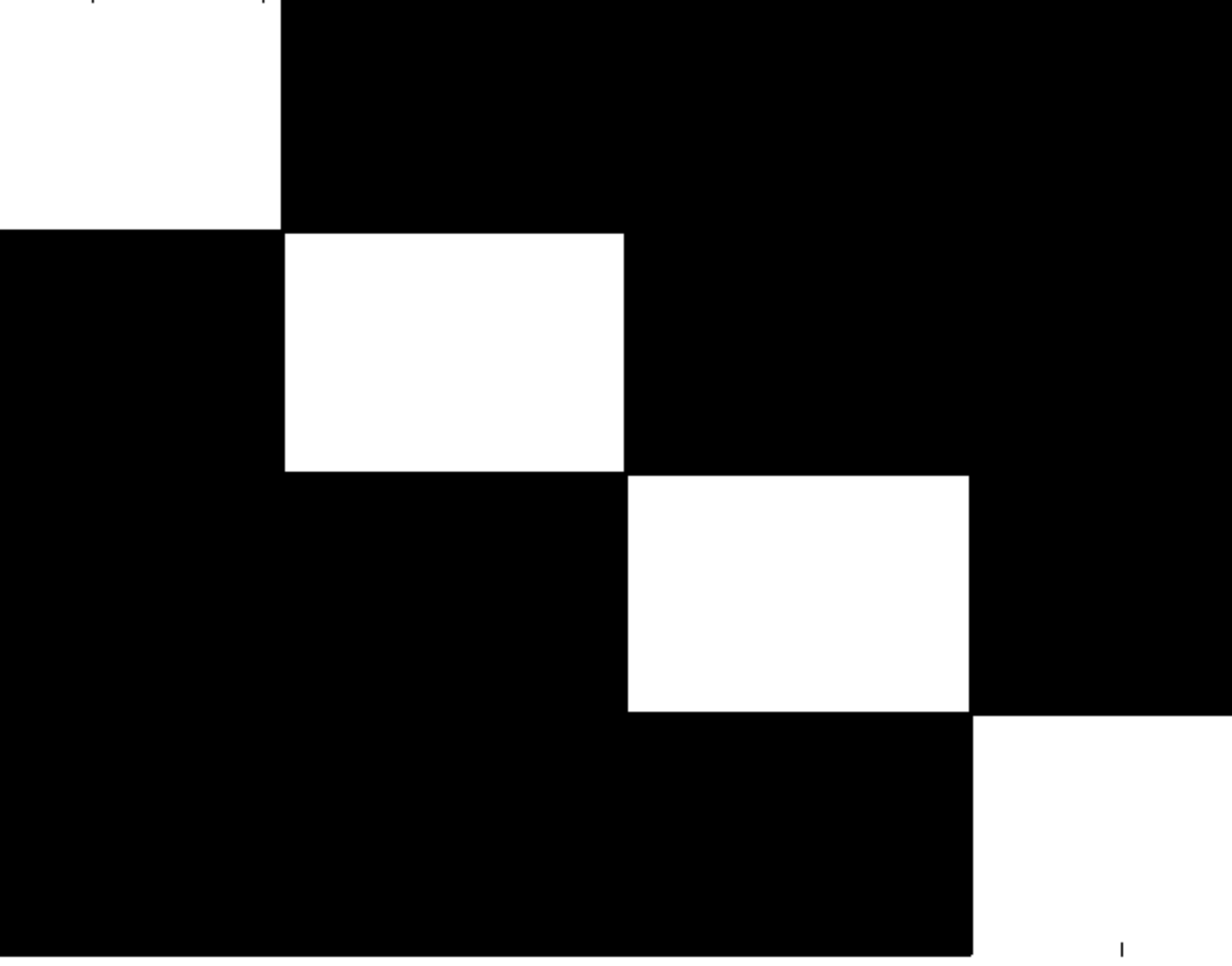}
			\label{fig:DisassortativeSBM}
			}
		\subfigure[{Core-Periphery}]{
			\includegraphics[width = .22\textwidth, clip = true, trim = 1.25in 3.1in 1in 3in]{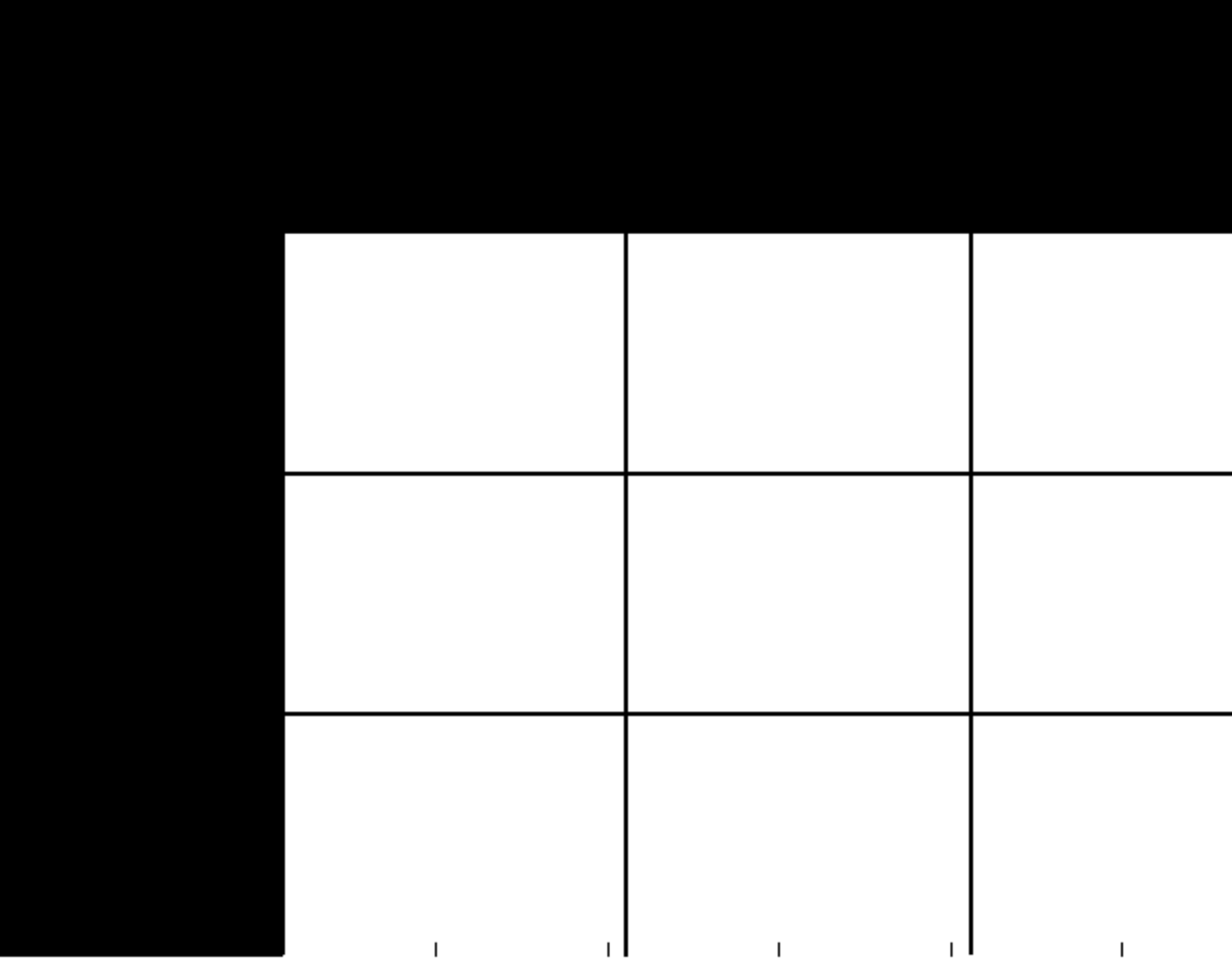}
			\label{fig:CorePeripherySBM}
			}
		\subfigure[{Ordered}]{
			\includegraphics[width = .22\textwidth, clip = true, trim = 1.25in 3.1in 1in 3in]{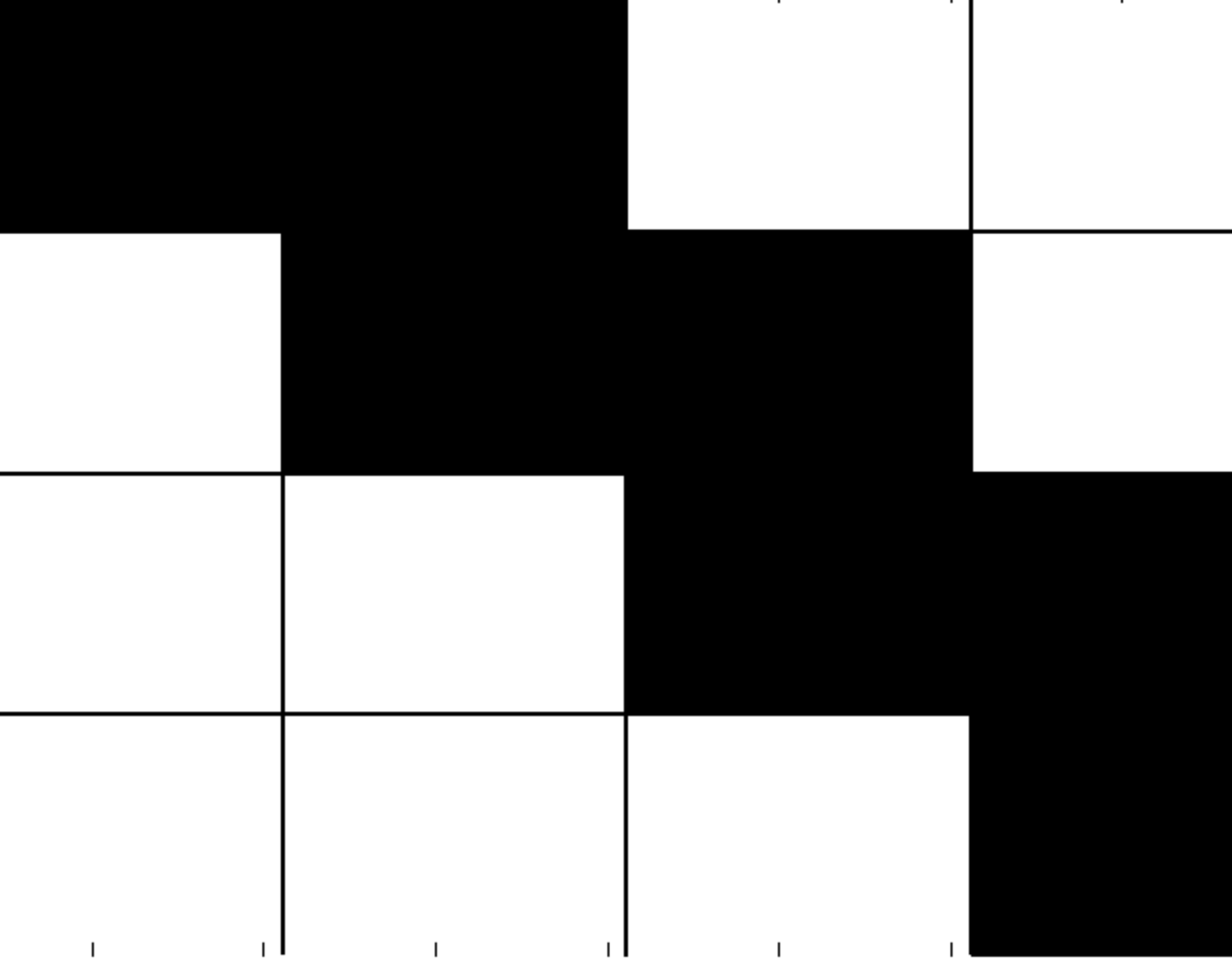}
			\label{fig:OrderedSBM}
			}
		
		\caption{
		Examples of structure that can be learned using the SBM. 
		The first row shows the abstract connections between four groups (blue, red, green, and purple).
		The second row shows the `block' structure found in the adjacency matrix after sorting by group membership; black corresponds to edges and white corresponds to non-edges.
		(a) Assortative structure: edges mainly exist within groups. 
		(b) Disassortative structure: edges mainly exist between distinct groups. 
		(c) Core-Periphery structure: the `core' (blue) connects mainly with itself and the `periphery' (red, green, and purple), while the `periphery' mainly connects with the `core'. 
		(d) Ordered structure: blue connects to red, red connects to green, and green connects to purple.
		}
		\label{fig:SBMExamples}
\end{figure} 
%

The stochastic block model (SBM)~\cite{holland:laskey:leinhardt:1983, wang:wong:1987} is a popular generative model for learning community structure in unweighted networks. 
In its classic form, the SBM is a probabilistic model of pairwise interactions among $n$ vertices. 
Each vertex $i$ belongs to one of $K$ latent groups or ``blocks'' denoted by $z_i$, and each edge $A_{ij}$ exists with a probability $\theta_{z_i z_j}$ that depends only on the group memberships of the connecting vertices. 
Vertices in the same block are stochastically equivalent, indicating their equivalent roles in generating the network's structure.
The SBM is fully specified by a vector $z$ denoting the group membership of each vertex and a $K\times K$ matrix $\theta$ of edge bundle probabilities, where $\theta_{k,k'}$ gives the probability that a vertex in group $k$ connects to some vertex of group $k'$. 

The SBM is popular in part because it can generate a wide variety of large-scale patterns of network connectivity depending on the choice of $\theta$ (Figs~\ref{fig:SBMExamples}(a-d)).
For example, if the diagonal elements of $\theta$ are greater than its off-diagonal elements, the block structure is assortative, with communities exhibiting greater edge densities within than between them (Fig.~\ref{fig:AssortativeSBM})---a common pattern in social networks~\cite{newman_mixing_2003}. 
Reversing the pattern in $\theta$ generates disassortative structure (Fig.~\ref{fig:DisassortativeSBM}), which is often found in language and ecological networks~\cite{newman_networks:_2010}. 
Other choices of $\theta$ can generate hierarchical, multi-partite, or core-periphery patterns ~\cite{clauset:moore:newman:2008,park_dynamic_2010}. 
The SBM also has been generalized for count-valued data, degree-correction~\cite{karrer_stochastic_2011}, bipartite structure~\cite{larremore_efficiently_2014}, and categorical values~\cite{guimera_network_2013}. 

In addition to this flexibility, the SBM's probabilistic structure provides a principled approach to quantifying uncertainty of group membership, an attractive feature in unsupervised network analysis.
This structure has led to theoretical guarantees, including consistency of the SBM estimators~\cite{celisse_consistency_2012} and the identifiability and consistency of latent block models~\cite{ambroise_new_2012,allman_parameter_2011}.

However, each of these models assumes an unweighted network, where edge presence or absence is represented as a binary variable (or perhaps a count-valued variable), while most real-world networks have weights, e.g., interaction frequency, volume, or character. Such information is typically discarded via thresholding before analysis, which can obscure or distort latent structure~\cite{thomas_valued_2011}.
\begin{figure}[tb]
\centering
		\subfigure[Example Network]{
	\raisebox{0.65in}{ 
	\includegraphics[width = .3\textwidth]{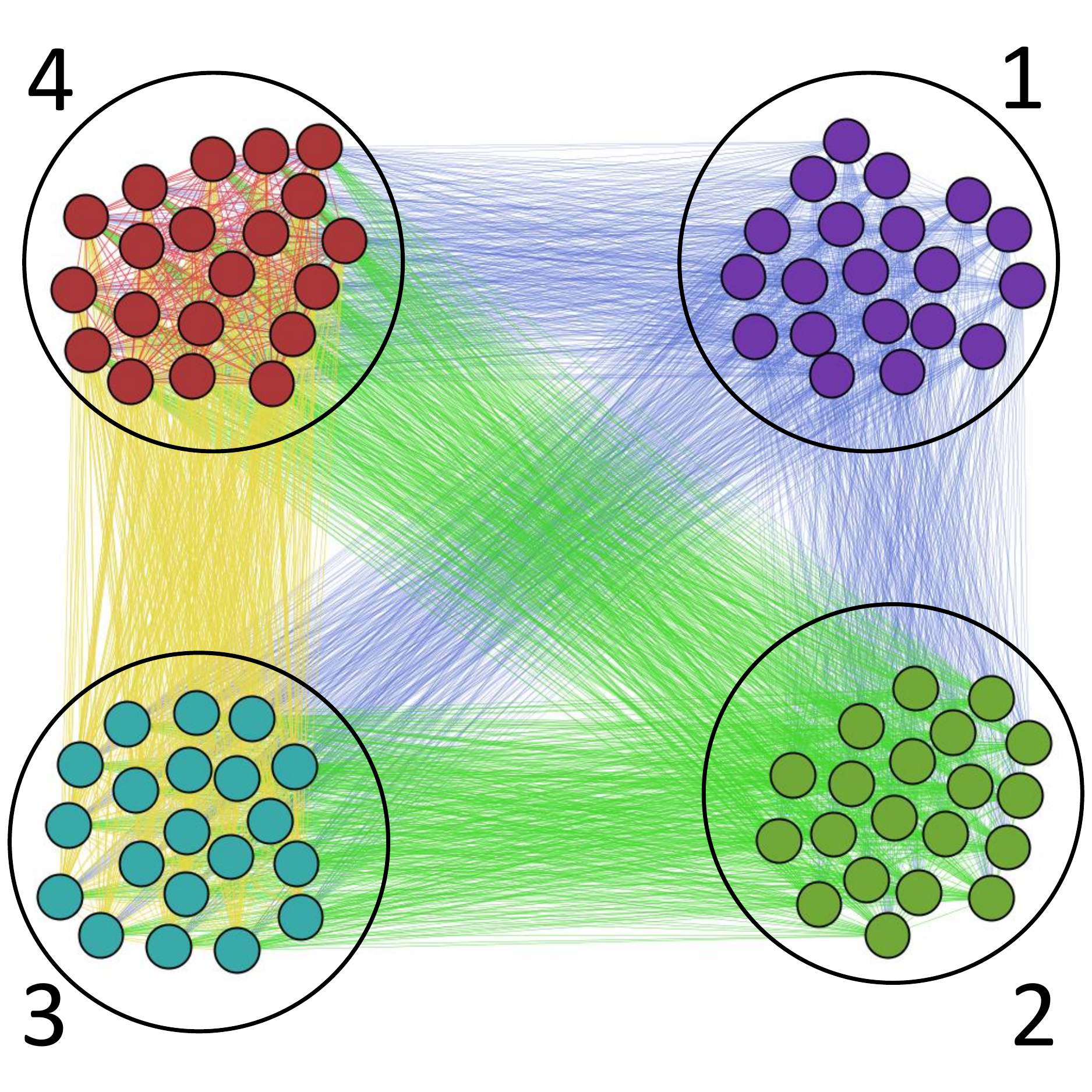}}
	}
		\subfigure[NMI vs Threshold]{
	\includegraphics[width = .55\textwidth]{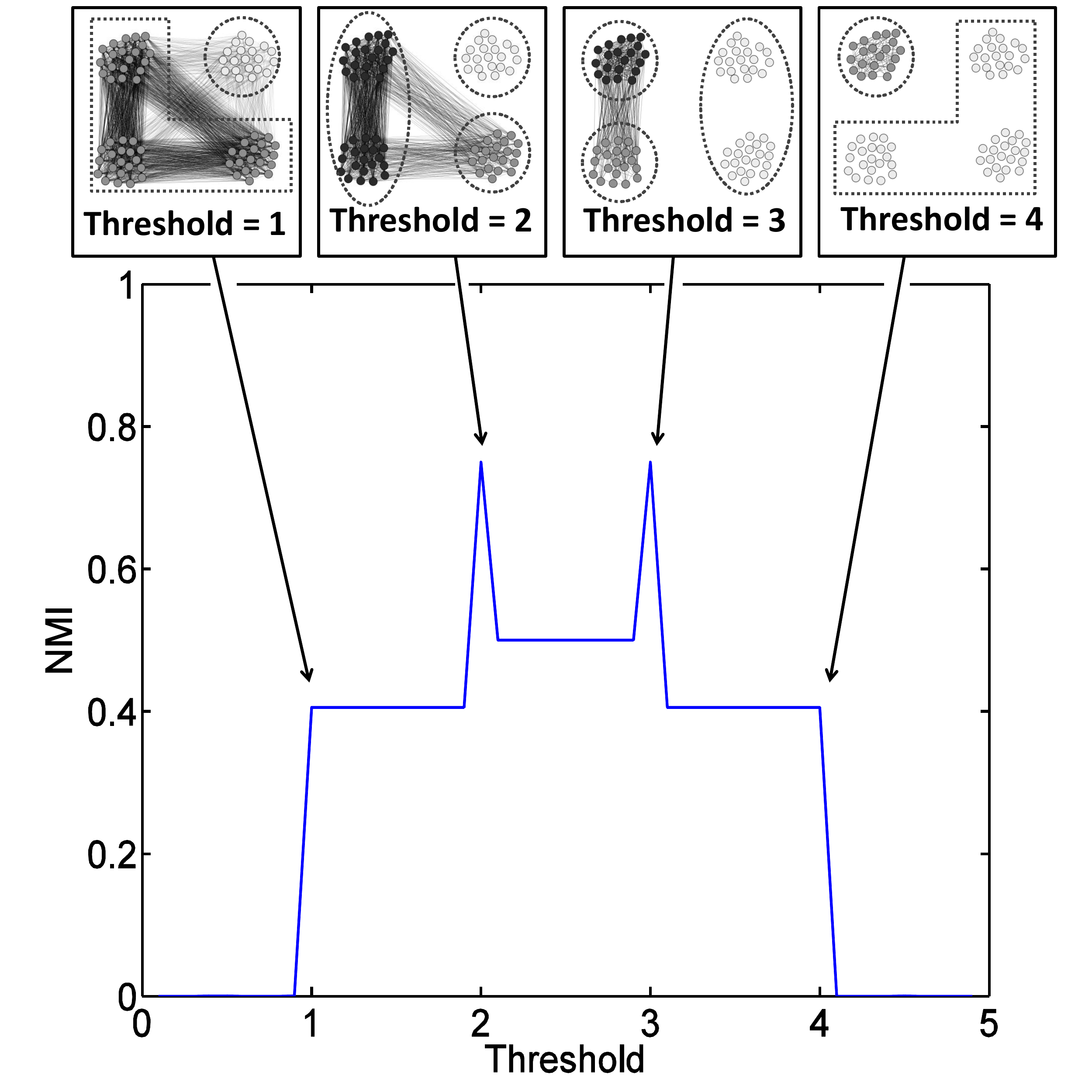}
	}
\caption{
    (a) An example of a weighted network where thresholding will never succeed.
    (b) A plot of the normalized mutual information (NMI) between the true community structure and inferred SBM community structure after thresholding at various threshold values (averaged over 100 trials). 
    Examples of community structure found by thresholding are shown above the graph (different colors represent different communities).
    As the NMI is less than $1$ for all threshold values, the SBM after thresholding never infers the true community structure shown in (a).
    }
\label{Fig:ThreshEx}
\end{figure}
To illustrate this loss of information from thresholding, consider a toy network of four equally-sized groups labeled 1--4 (see Fig.~\ref{Fig:ThreshEx}), where each edge $(i,j)$ is assigned a weight equal to the smaller of the endpoints' group labels, plus a small amount of noise. Edges between groups are thus assigned weights near 1, 2, or 3, while those within a group are assigned weights near 1--4. This model is obviously unrealistic, but serves to illustrate the common consequences of applying a global threshold to edge-weighted networks.

To apply the SBM to this simple network, we must convert it into an unweighted network by discarding edges with weights less than some threshold. To illustrate the results of this action, we consider all possible thresholds, and compute the average normalized mutual information (NMI) between the best community structure found using the SBM and the true structure (Fig.~\ref{Fig:ThreshEx}).
No matter what threshold we choose, edges are divided into at most three groups: those with weight above, at, or below the threshold. 
The SBM can thus recover a maximum of three groups, rather than the four planted in this network, and the threshold determines which three groups it finds. No threshold yields the correct inference here, because thresholding discards edge weight information. 

Instead of thresholding, we could use more complex methods, such as using multiple thresholds or a binning scheme, to convert a weighted network into an unweighted or count-valued network of some sort. 
These methods would perform better than applying a single threshold, at the cost of additional complexity in specifying multiple threshold or bin values. 
Regardless of the method, these approaches will still discard potentially useful edge weight information.
To exploit the maximal amount of information in the original data in recovering the true hidden structure, we should prefer to model the edge weights directly.

In this paper, we introduce the \emph{weighted stochastic block model} (WSBM), a generalization of the SBM that can learn from both the presence and weight of edges.   
The weighted stochastic block model provides a natural solution to this problem by generalizing the SBM to learn from both types of edge information.
Specifically, the WSBM models each weighted edge $A_{ij}$ as a draw from a parametric exponential family distribution, whose parameters depend only on the group memberships of the connecting vertices $i$ and $j$. 
It includes as special cases most standard distributional forms, e.g., the normal, the exponential, and their generalizations, and enables the direct use of weighted edges in recovering latent group or block structure.
This paper generalizes and extends our previous work \cite{aicher_adapting_2013}.

We first describe the form of the WSBM, which combines edge existence and weight information. 
We then derive a variational Bayes algorithm for efficiently learning WSBM parameters from data. 
Applying this algorithm to a small real-world weighted network, we show that the SBM and WSBM can learn distinct latent structures as a result of observing or ignoring edge weights. 
Finally, we compare the performance of the WSBM to alternative methods for two edge prediction tasks, using a set of real-world networks.
In all cases, the WSBM performs as well as alternatives on edge-existence prediction, and outperforms all alternatives on edge-weight prediction.
This model thus enables the discovery of latent group structures in a wider range of networks than was previously possible.

\section{Weighted Stochastic Block Model}
We begin by reviewing the SBM and exponential families, and then describe a natural generalization of the SBM to weighted networks.
In what follows, we consider the general case of directed graphs; undirected graphs are a special case of this model.

In the SBM, the network's adjacency matrix $A$ contains binary values representing edge existences, i.e., \mbox{$A_{ij}\! \in\! \{0,1\}$}, the integer $K$ denotes a fixed number of latent groups, and the vector $z$ contains the group label of each vertex $z_i\! \in\! \{1,\ldots,K\}$.
The number of latent groups $K$ controls the model's complexity and may be chosen in a variety of ways---we defer a discussion of this matter until section~\ref{sec:selecting:k}.
Each possible group assignment vector $z$ represents a different partition of the vertices into $K$ groups, and each pair of groups $(k k')$ defines a ``bundle'' of edges that run between them. 
The SBM assigns an edge existence parameter to each edge bundle $\theta_{k k'}$, which we represent collectively by the $K$-by-$K$ matrix $\theta$. 
The existence probability of an edge $A_{ij}$ is given by the parameter $\theta_{z_i z_j}$ that depends only on the group memberships of vertices $i$ and $j$. 

Assuming that each edge existence $A_{ij}$ is conditionally independent given $z$ and $\theta$, the SBM's likelihood function is
\begin{equation}
\label{eq:SBM}
\Pr(A \, | \, z,\theta) = \prod_{ij} \theta_{z_i z_j}^{A_{ij}} \left(1-\theta_{z_i z_j}\right)^{1-A_{ij}} \enspace ,
\end{equation}
which we may rewrite as 
\begin{equation}
\Pr(A \, | \, z,\theta) = \prod_{ij} \exp\!\left( A_{ij} \cdot \log\!\left(\frac{\theta_{z_i z_j}}{1-\theta_{z_i z_j}}\right) + \log\left(1-\theta_{z_i z_j}\right) \right) \enspace . \nonumber
\end{equation}
Thus, the likelihood has the form of an exponential family
\begin{equation}
\label{eq:universal}
\Pr(A \,|\, z,\theta) \propto \exp\!\left( \sum_{ij} T(A_{ij}) \cdot \eta(\theta_{z_i z_j}) \right) \enspace,
\end{equation}
where $T(x) = (x,1)$ is the vector-valued function of sufficient statistics of the Bernoulli random variable and $\eta(x) = \left(\log[x/(1-x)],\log[1-x]\right)$ is the vector-valued function of natural parameters.  Appendix \ref{app:expfamily} provides further details about exponential families.

\begin{figure}[tb]
\centering
	\includegraphics[height = 0.18\textheight, clip = true, trim = 0in 0in 0in 0.0in]{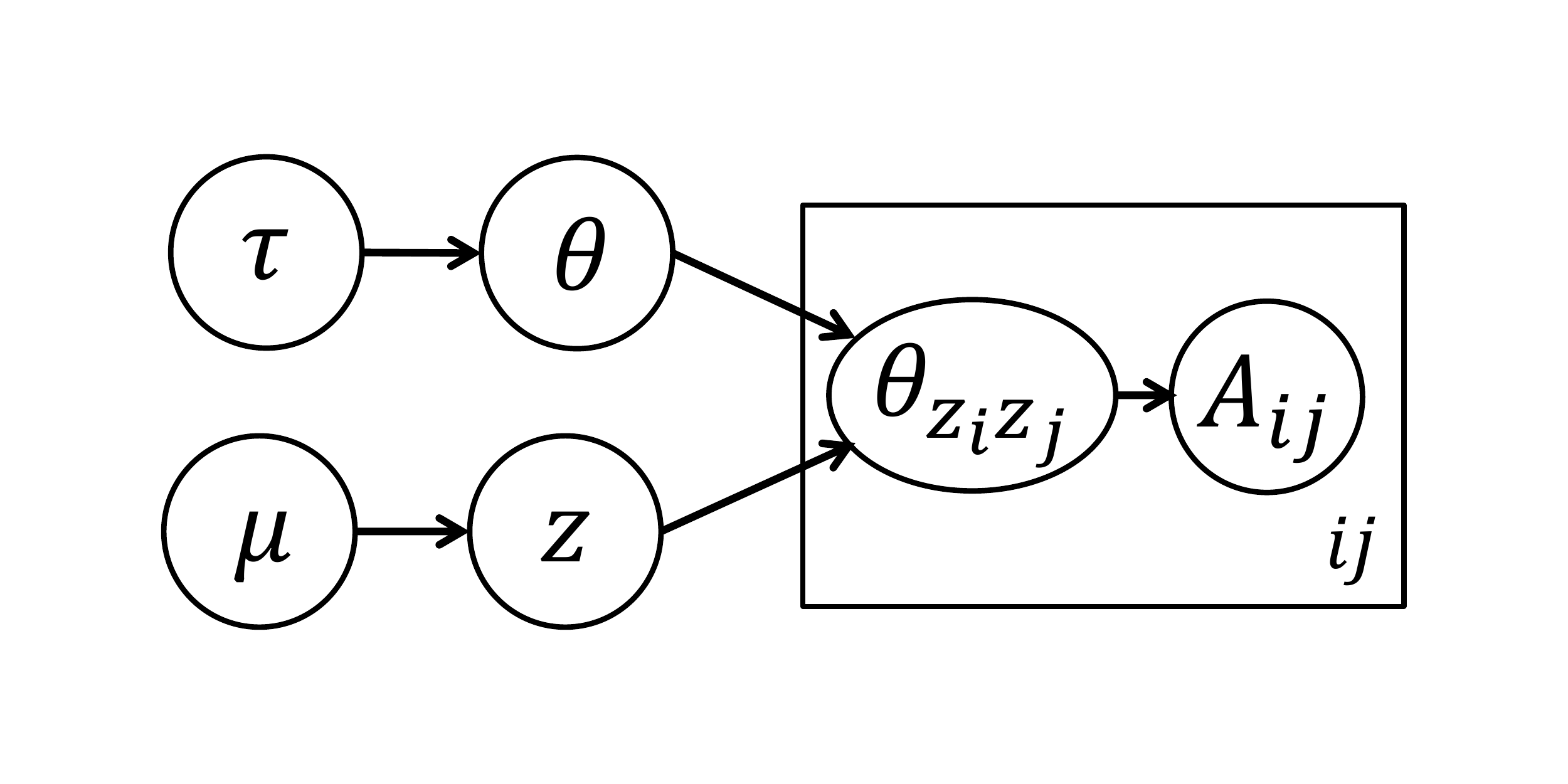}
\caption{
    Graphical model for the WSBM. Each weighted edge $A_{ij}$ (plate) is distributed according to the appropriate edge parameter $\theta_{z_i,z_j}$ for each observed interaction $(i,j)$. 
		In our variational Bayes inference scheme, the WSBM's latent parameters $z,\theta$ are themselves modeled as random variables distributed according to $\mu,\tau$, respectively.
		We highlight that the arrow from $z$ to $\theta_{z_i,z_j}$ hides the complex relational structure between each $z_i$. 
    }
\label{Fig:GraphicalModel}
\end{figure}

This choice of functions $(T,\eta)$ produces binary-valued edge weights. 
By choosing an appropriate but different pair of functions $(T,\eta)$, defined on some domain $\mathcal{X}$ and $\mathcal{\Theta}$ respectively, we may specify a stochastic block model whose weights are drawn from an exponential family distribution over $\mathcal{X}$. 
As in the SBM, this weighted stochastic block model (WSBM) is defined by a vector $z$ and matrix $\theta$, but now each $\theta_{z_{i}z_{j}}$ specifies the parameters governing the weight distribution of the $(z_{i}z_{j})$ edge bundle. 
Figure~\ref{Fig:GraphicalModel} visualizes the dependencies in the WSBM's likelihood function as a graphical model.

The generative process of creating a weighted network from the WSBM consists of the following steps.
\begin{itemize}[noitemsep]
\item For each vertex $i$, assign a group membership $z_i$.
\item For each pair of groups $(k,k')$, assign an edge bundle parameter $\theta_{k k'} \in \mathcal{\Theta}$
\item For each edge $(i,j)$, draw $A_{ij} \in \mathcal{X}$ from the exponential family $(T,\eta)$ parametrized by $\theta_{z_i z_j}$.
\end{itemize}
The community structure of the WSBM retains the stochastic equivalence principle of the classic SBM, in which all vertices in a group maintain the same probabilistic connectivity to the rest of the network. 

For example, if the edge weights are real-valued $\mathcal{X} = \mathbb{R}$, then we may choose to model the edge weights with the normal distribution, which has sufficient statistics $T \!=\! (x,x^2,1)$ and natural parameters $\eta = (\mu/\sigma^2, -1/(2\sigma^2),-\mu^2/(2\sigma^2))$. 
Instead of edge-existence probabilities, each edge-bundle $(z_{i} z_{j})$ is now parameterized by a mean and variance $\theta_{z_{i} z_{j}} = (\mu_{z_{i} z_{j}}, \sigma^2_{z_{i} z_{j}})$.
In this case, the likelihood function would be
\begin{equation}
\Pr(A\,|\,z,\mu,\sigma^2) = \prod_{ij} \mathcal{N}\left(A_{ij} \, | \, \mu_{z_i z_j} , \sigma^2_{z_i z_j}\right) = \prod_{ij} \exp\left( 
A_{ij} \cdot \frac{\mu_{z_i z_j}}{\sigma^2_{z_i z_j}} - A_{ij}^2 \cdot \frac{1}{2 \sigma^2_{z_i z_j}} - 1 \cdot \frac{\mu^2_{z_i z_j}}{\sigma^2_{z_i z_j}}  \right)  \enspace. 
\end{equation}
That is, this particular WSBM uses a normal distribution instead of a Bernoulli distribution to model the values observed in an edge bundle. 
We emphasize that the choice of the normal distribution is merely illustrative. 

This construction produces complete graphs, in which every pair of vertices is connected by an edge with some real-valued weight. 
For a complete network, this formulation may be entirely sufficient. 
However, most real-world networks are sparse, with only $O(n)$ pairs having a connection that may have a weight, and a dense model like this one cannot be applied directly. 
We now describe how sparsity can be naturally incorporated within our model, which also produces more scalable inference algorithms.

\subsection{Sparse Weighted Graphs}
\label{sec:swg}


A key insight for modeling edge-weighted sparse networks lays in clarifying the meaning of zeros in a weighted adjacency matrix. Typically, a value $A_{ij}\!=\!0$ may represent one of three things: (i) the absence of an edge, (ii) an edge that exists but has weight zero, or (iii) missing data, i.e., an unobserved interaction. 
In both of the former two cases, we do in fact observe the interaction, while in the latter, we do not. For observed interactions, we call the observed non-interaction to be a ``non-edge,''
and we let $A_{ij}\!=\!0$ denote the presence of an edge with weight zero.
In many empirical networks, distinct types of interactions may have been confounded, e.g., non-edges, edges with zero weight, and unobserved interactions may all be assigned a value $A_{ij}\!=\!0$. However, for accurate inference, this distinction can be important.
For example, a non-edge may indicate an interaction that is impossible to measure, which is distinct from choosing not to measure the interaction (an unobserved interaction) or an interaction with weight zero.  

Here, we assume that these three types of interactions are distinguished in our input data. 
This creates two types of information: information from edge existence (non-edges vs weighted edges) and information from edge weight (the weighted values). 
To handle these two types of information, the WSBM then models an edge's existence as a Bernoulli or binary random variable, as in the SBM, and models an edge's weight using an exponential family distribution. Terms corresponding to unobserved interactions contribute no information to inference and are dropped from the likelihood function.
If the pair $(T_e,\eta_e)$ denotes the family of edge-existence distributions and the pair $(T_w,\eta_w)$ denotes the family of edge-weight distributions then we may combine their contributions in the likelihood function via a simple tuning parameter $\alpha \!\in\! [0,1]$ that determines their relative importance in inference
\begin{equation}
\label{eq:general}
\log\Pr(A\, |\,z,\theta) = \alpha \sum_{ij \in E} T_e(A_{ij}) \cdot \eta_e\!\left(\theta^{(e)}_{z_i z_j}\right) + (1-\alpha) \sum_{ij \in W} T_w(A_{ij}) \cdot \eta_w\!\left(\theta^{(w)}_{z_i z_j}\right) \enspace,   
\end{equation}
where $E$ is the set of observed interactions (including non-edges) and $W$ is the set of weighted edges ($W \subset E$).  
This generalization 
can be reduced to the compact form of Eq.~\eqref{eq:universal} by combining the vectors $\alpha T_e$ with $(1-\alpha)T_w$ and $\eta_e$ with $\eta_w$.

By tuning $\alpha$, we can learn different latent structures.
When $\alpha \!=\! 1$, the model ignores edge weight information and reduces to the SBM.
When $\alpha \!=\! 0$, the model treats edge absence as if it were unobserved, and fits only to the weight information.
When $0\!<\!\alpha\!<\!1$, the likelihood combines information from both edge existence and weights.
In principle, the best choice of $\alpha$ could also be learned, but we leave this subtle problem for future work. 
In practice, we often find that $\alpha=1/2$, giving equal weight to both types of information, works well.

\subsection{Degree Correction}
The last piece of the WSBM is a generalization to naturally handle heavy-tailed degree distributions, which are ubiquitous in real-world networks and are known to cause the SBM to produce undesirable results, e.g., placing all high-degree vertices in a group together, regardless of their natural community membership~\cite{karrer_stochastic_2011}.

Karrer and Newman introduced an elegant extension of the SBM that circumvents this behavior. 
In their ``degree corrected'' SBM (here DCBM), they add vertex degree information into the generative model by adding an ``edge-propensity'' parameter $\phi_i$ to each vertex~\cite{karrer_stochastic_2011}.
As a result, the number of edges that exist between a pair of vertices $i$ and $j$ is a Poisson random variable with mean $\phi_i \phi_j \theta_{z_i,z_j}$.
Because vertices with high propensity are more likely to connect than vertices with low propensity, the propensity parameters $\phi$ allow for heterogenous degree distributions within groups.
In the DCBM, vertices in the same block are no longer stochastically equivalent, but have similar group-level connectivity patterns conditioned on their propensity parameters $\phi$.  

The likelihood function for this model is
\begin{equation*}
\Pr(A \,|\, z,\theta,\phi) \propto \prod_{ij} \left(\phi_i \phi_j \theta_{z_i z_j} \right)^{A_{ij}} \exp\left(-\phi_i \phi_j \theta_{z_i z_j} \right) \enspace,
\end{equation*}
where the maximum likelihood estimate of each propensity parameter $\phi_i$ is simply the vertex degree $d_i$~\cite{karrer_stochastic_2011}. 
By fixing $\phi_i=d_i$, we can rewrite the DCBM in the exponential family form
\begin{equation}
\Pr(A \,|\, z,\theta,\phi) \propto \prod_{ij} \exp\!\left( 
A_{ij} \cdot \log \theta_{z_i z_j} -d_i d_j \cdot \theta_{z_i z_j}
\right)
 \enspace, \label{eq:dcbm}
\end{equation}
where the sufficient statistics are $T \!=\! (A_{ij}, -d_i d_j)$ and the natural parameters are $\eta \!=\! (\log\theta_{z_i z_j}, \theta_{z_i z_j})$. 
Thus, to derive a degree-corrected weighted stochastic block model, we simply replace the SBM contribution in Eq.~\eqref{eq:general} with that of the DCBM in Eq.~\eqref{eq:dcbm}. We note that this model can easily extended to included in- and out-propensity parameters for directed networks.

This degree-corrected weighted stochastic block model allows for heterogeneous degree distributions within groups by modeling vertex degree or rather the sum of edge existences.
This is distinct from what one might call a `strength'-corrected SBM that produces heterogeneous weight distributions within edge bundles by modeling vertex strength (the sum of a vertex's edge weights).
This `strength'-corrected model is not consider here and is an area for future work.
  
\section{Learning Latent Block Structure}
Given some sparse weighted graph $A$, we recover the underlying communities by learning the parameters $z,\theta$. Any of a large number of standard approaches can be used to optimize the likelihood function for the WSBM. Here, we describe an efficient variational Bayes approach~\cite{attias_variational_2000, hofman_bayesian_2008}, which effectively handles one technical difficulty in fitting the model to real data.

Specifically, learning the parameters $z,\theta$ by directly maximizing the likelihood in Eq.~\eqref{eq:universal} can suffer degenerate solutions under continuous valued weights.
For instance, consider the WSBM with normally distributed edge weights, where some bundle of edges has all-equal weights. In this case, the maximum likelihood estimate is a variance parameter equal to zero, which creates a degeneracy in the likelihood calculation. 
This case is not pathological, as a poor choice of partition $z$---chosen, perhaps, inadvertently over the course of maximizing the likelihood---can easily create two small groups with only a few edges, each with the same weight, between them.
This problem has not previously been identified in the block-modeling literature because the SBM is a model where edge ``weights'' are discrete Bernoulli random variables, whose parameters are never degenerate.

We solve this problem using Bayesian regularization. 
In the Bayesian framework, we treat the parameters as random variables and assign an appropriate prior distribution $\pi$ to our parameters $z,\theta$.
If we treat the prior distribution as the probability of the parameters $\pi(z,\theta) = \Pr(z,\theta)$ then we may calculate the posterior distribution as the probability of the parameters conditioned on the data $\pi^*(z,\theta) = \Pr(z,\theta | A)$ through Bayes' law
\begin{equation*}
\pi^*(z,\theta) \propto \Pr(A|z,\theta)\pi(z,\theta) \enspace.
\end{equation*}
After calculating the posterior distribution, we may either return our posterior beliefs $\pi^*$ about the parameters $z,\theta$ or further calculate a point estimate to minimize a posterior expected loss with respect to a given loss function~\cite{ohagan_kendalls_2009,robert_bayesian_2007}.
In both cases, it suffices to calculate the posterior $\pi^*$.
The maximum likelihood estimate corresponds to only maximizing the likelihood $\Pr(A|z,\theta)$.
The inclusion of the prior distribution $\pi$ prevents the posterior distribution $\pi^*$ from over-fitting to the degenerate maximum likelihood solution and therefore estimation can proceed smoothly.

However, the posterior distribution is generally difficult to calculate analytically. Instead, we approximate $\pi^{*}(z,\theta)$ by a factorizable distribution $q(z,\theta) = q_z(z)q_\theta(\theta)$, a common approach in both machine learning and statistical physics.
We select our approximation $q$ by minimizing its Kullback-Leibler (KL) divergence to the posterior 
\begin{equation*}
D_{\text{KL}}(q\, || \, \pi^{*}) = -\int q\, \log \frac{\pi^{*}}{q} \enspace.
\end{equation*}
The Kullback-Leibler divergence is a non-symmetric, non-negative, information-theoretic measure of difference between two distribution.
Thus, our approximation $q$ can be thought of as the closest approximation to the posterior $\pi^*$, subject to factorization and distribution constraints. 

Expanding the constant likelihood $\log\Pr(A)$, we observe that minimizing the KL-divergence is equivalent to maximizing the functional $\G(q)$ defined as follows. Let
\begin{align*}
\log \Pr(A) &= \int_\Theta \sum_{z \in Z} q(z,\theta) \ d\theta \ \log \Pr(A) \\
&= \int_\Theta \sum_{z \in Z} q(z,\theta) \log\frac{\Pr(A , z , \theta)}{ \Pr(z, \theta | A)} \ d\theta \\
&= \int_\Theta \sum_{z \in Z} q(z,\theta) \log\frac{\Pr(A, z , \theta)}{q(z,\theta)} \ d\theta - \int_\Theta \sum_{z \in Z} q(z,\theta) \log\frac{\Pr(z , \theta | A)}{q(z,\theta)} \ d\theta \\
& = \G(q) + D_{KL}\left( q(z,\theta) || \pi^*(z,\theta)\right) \enspace,
\end{align*}
where
\begin{equation}
\G(q) = \int_\Theta \sum_{z \in Z} q(z,\theta) \log\frac{\Pr(A, z , \theta)}{q(z,\theta)} \ d\theta = \mathbb{E}_{q}\!\left(\log\Pr(A\, |\, z,\theta)\right) + \mathbb{E}_{q}\!\left(\log \frac{\pi( z , \theta)}{q( z , \theta)}\right) \enspace . \label{eq:Gp}
\end{equation}
The first term of Eq.~\eqref{eq:Gp} is the expected log-likelihood under the approximation $q$ and the second term is the negative KL-divergence of the approximation $q$ from the prior $\pi$.
Therefore, we aim to maximize the expected log-likelihood of the data and weakly constrain the approximation to be close to the prior. The second term serves as a regularizer which prevents over-fitting and eliminates the aforementioned maximum likelihood degeneracies. 
In practice, the first term dominates the second term given sufficient data and approximates the maximum likelihood estimation.

Because the KL-divergence is non-negative, we can think of $\G(q)$ as a functional lower bound on the log-evidence or marginal log-likelihood, that is,
%
\begin{equation}
\label{eq:lowerboundG}
\log \Pr(A) = \G(q) + D_{\text{KL}}\left( q \,||\, \pi^{*}\right) \geq \G(q) \enspace.
\end{equation}
Maximizing $\G(q)$ is equivalent to minimizing the KL divergence $D_{\text{KL}}(q\, || \,\pi^{*})$ because the log-evidence $\log \Pr(A)$ is constant.
Therefore as we maximize $\G(q)$, our approximation $q$ gets closer to the true posterior $\pi^{*}$.
For more details on variational Bayesian inference in graphical models, we refer the interested reader to Ref.~\cite{attias_variational_2000}. 

\subsection{Conjugate Distributions}
To calculate $\G$ in practice, we must assign prior distributions $\pi$ to our parameters and place constraints on the distributions of our approximation $q$.
For mathematical convenience, we choose $\pi$ and restrict $q$ to be the product of parameterized conjugate distributions. 
Because $q$ takes a parameterized form, maximizing the functional $\G(q)$ over all factorized distributions $q$ simplifies to maximizing $\G(q)$ over the parameters of $q$. 

For the edge bundle parameters $\theta$, the standard conjugate prior of the parameter of an exponential family $(T,\eta)$ is
\begin{equation}
\pi(\theta) = \frac{1}{Z(\tau)}\exp\left(\tau \cdot \eta(\theta)\right) \enspace ,
\end{equation}
where $\tau$ parameterizes the prior and $Z(\tau)$ is a normalizing constant for fixed $\tau$. 

For notational convenience, we let $r$ index into the $K \times K$ edge-bundles between groups; hence $\theta = (\theta_1,...,\theta_r)$. 
When we update the prior based on the observed weights in a given edge bundle $r$, the posterior's parameter becomes $\tau^* = \tau + T_r$, where $T_r$ is the sufficient statistic of the observed edges. 
Thus $\tau$ can be viewed as a set of pseudo-observations that push the likelihood function away from the degenerate cases so that every edge bundle, no matter how small or uniform, produces a valid parameter estimate.

For the vertex labels $z$, the natural conjugate prior is a categorical distribution with parameter $\mu \in \mathbb{R}^{n\times k}$.
The parameter $\mu_i(k)$ represents the probability that vertex $i$ belongs to group $k$ in all of its interactions. 
If the probability in parameter $\mu_i$ is spread among multiple groups, then this indicates uncertainty in the membership of vertex $i$ and not mixed membership.
We fit $\mu_i$ directly, with flat prior $\mu_0(k) = 1/K$. 

The form of our prior is thus
\begin{equation}
\pi(z, \theta \,|\, \mu_0, \tau_0) = \prod_i \mu_0(z_i) \times \prod_{r}  \frac{1}{Z(\tau_0)}\exp\!\left(\tau_0 \cdot \eta(\theta_r)\right)  \enspace ,
\end{equation}
where $\mu_0$, $\tau_0$ are the parameters for the priors $\pi_i$, $\pi_r$, picked to be a ``non-informative'' reference prior~\cite{berger_development_1992} or flat. 

Similarly, our approximation $q$ takes the form
\begin{equation}
q(z, \theta \,|\, \mu, \tau) = \prod_i \mu_{i}(z_i) \times\prod_{r}  \frac{1}{Z(\tau_r)}\exp\!\left(\tau_r \cdot \eta(\theta_r)\right)\enspace .
\end{equation}

\subsection{An efficient algorithm for optimizing $\G$} 
Now we consider maximizing $\G$ over $q$'s parameters $\mu_i$, $\tau_r$. 
To simplify notation, let $\left\langle T \right\rangle_r$, $\left\langle \eta \right\rangle_r$ be the expected values of the sufficient statistics $T_r$ and natural parameters $\eta_r$ under the approximation $q$, that is, we set
\begin{align}
  \left\langle T \right\rangle_r & = \sum_{ij} \sum_{(z_i,z_j) = r} \mu_i(z_i) \, \mu_j(z_j) \, T(A_{ij}) \\
  \left\langle \eta \right\rangle_r &= \left. \frac{\partial}{\partial \tau} \log Z(\tau) \right|_{\tau=\tau_r} \enspace . 
\end{align}

Substituting the conjugate prior forms of $\pi$, $q$ into $\G$ thus yields
\begin{equation} \G \propto \sum_r \left(\left\langle T \right\rangle_r + \tau_0 - \tau_r\right) \cdot \left\langle \eta \right\rangle_r + \sum_r \log \frac{Z(\tau_r)}{Z(\tau_0)} + \sum_i \sum_{z_i} \mu_i (z_i) \log \frac{\mu_0 (z_i)}{\mu_i (z_i)} \enspace .
\end{equation} 

To optimize $\G$, we take derivatives with respect to $q$'s parameters $\mu$, $\tau$ and set them to zero. 
We iteratively solve for the maximum by updating $\mu$ and $\tau$ independently.

For the edge bundle parameters $\tau$, the derivative of $\G$ is
\begin{align}
\frac{\partial \G}{\partial \tau_r} &= \left(\left\langle T \right\rangle_r + \tau_0 - \tau_r\right) \frac{\partial \left\langle \eta \right\rangle_r}{\partial \tau_r} \enspace ,
\end{align}
and setting this equal to zero yields a compact update equation
\begin{equation}
\label{eq:tauupdate}
\tau_r = \tau_0+ \left\langle T \right\rangle_r \enspace
\end{equation}
for each edge bundle $r$.

For the vertex label parameters $\mu$, we include Lagrange multipliers $\lambda_i$ to enforce the constraint $\sum_{z} \mu_i(z) = 1$. 
Setting the derivative of $\G$ with respect to $\mu_i$ equal to $\lambda_i$ yields
\begin{equation*}
\frac{\partial \G}{\partial \mu_i(z)} = \sum_r \left( \frac{\partial \left\langle T \right\rangle_r}{\partial \mu_i(z)} \cdot \left\langle \eta \right\rangle_r \right) - \log \mu_i(z) = \lambda_i \enspace ,
\end{equation*}
where
\begin{equation*}
\frac{\partial \left\langle T \right\rangle_r}{\partial \mu_i(z)} := \sum_{z' : (z,z') = r} \,\,\, \sum_{j \neq i} T(A_{ij}) \mu_j(z') \enspace .
\end{equation*}
Solving for $\mu_i(z)$ yields a compact update equation
\begin{equation}
\label{eq:muupdate}
\mu_i(z) \propto \exp\! \left( \sum_r \frac{\partial \left\langle T \right\rangle_r}{\partial \mu_i(z)} \cdot \left\langle \eta \right\rangle_r \right) \enspace ,
\end{equation}
where each $\mu_i$ is normalized to a probability distribution. 
To calculate the $\mu_i$ values, we iteratively update each $\mu_i$ from some initial guess until convergence to within some numerical tolerance.

\begin{algorithm}[tb]
   \caption{Variational Bayes for WSBM}
   \label{alg:VB}
\begin{algorithmic}
   \STATE {\bfseries Input:} Edge-weighted network $A$ and Model $\M$
   \STATE Initialize $\mu$
   \REPEAT
   \FORALL{$r = 1,\ldots, K^2$}
   \STATE Set $\left\langle T \right\rangle_r := \sum_{ij} \sum_{(z_i,z_j) = r} \mu_i(z_i) \mu_j(z_j) T(A_{ij})$
   \STATE Set $\tau_r := \tau_0 + \left\langle T \right\rangle_r$
   \STATE Set $\left\langle \eta \right\rangle_r := \left. \frac{\partial}{\partial \tau} \log Z(\tau) \right|_{\tau=\tau_r}$ 
   \ENDFOR
   \REPEAT
   \FORALL{$i = 1,\ldots, n$}
   \STATE $\frac{\partial \left\langle T \right\rangle_r}{\partial \mu_i(z)} := \sum_{(z,z') = r} \sum_{j \neq i} T(A_{ij}) \mu_j(z') $
   \STATE $ \mu_i(z) \propto \exp \left( \sum_r \frac{\partial \left\langle T \right\rangle_r}{\partial \mu_i(z)} \cdot \left\langle \eta \right\rangle_r \right) $
   \ENDFOR 
   \UNTIL{$\mu$ converge}
   \UNTIL{$\mu,\tau$ converge }
   \RETURN $\mu,\tau$
\end{algorithmic}
\end{algorithm}

Algorithm~\ref{alg:VB} gives pseudocode for the full variational Bayes algorithm, which alternates between updating the edge-bundle parameters and the vertex label parameters using update equations Eqs. (\ref{eq:tauupdate}, \ref{eq:muupdate}). 
Updating $\theta$ is relatively fast. First, we calculate $\left\langle T \right\rangle_r$ and $\tau_r$ for each edge bundle $r$ and then update each $\left\langle \eta \right\rangle_r$, which takes $O(nK^2)$ time.
Updating $\mu$ is the limiting step of the calculation, as we iteratively update $\mu$ until convergence while holding $\theta$ fixed. 
To calculate $\partial \left\langle T \right\rangle_r/\partial \mu_i(z)$, each vertex must sum over its connected edges for each edge bundle, which takes $O(d_i K^2)$ time. 
If $m$ is the total number of edges in the network, then updating $\mu$ takes $O(mK^2)$ time.
In particular, if the total number of edges in the network is sparse $m = O(n)$, then updating $\mu$ takes $O(nK^2)$ time.

In practice, we would run the algorithm to convergence from a number of randomly-chosen initial conditions, and then select the best $\mu,\tau$. 

In addition to the variational Bayes algorithm above, we derive in Appendix~\ref{app:bp} an efficient loopy belief propagation algorithm~\cite{yedidia_understanding_2003, decelle_phasetransition_2011, yan_model_2012} for the WSBM on sparse graphs. The loopy belief propagation algorithm creates a more flexible approximation to the posterior distribution than the variational Bayes algorithm, but with a slightly higher computational cost. Small modifications for dealing with sparse weighted networks, are  described in Appendix~\ref{app:sparseweighted}. Finally, Appendix~\ref{app:code} describes how to obtain our implementation of these methods.

\subsection{Selecting $K$ with Bayes factors}
\label{sec:selecting:k}
As with most stochastic block models, the number of groups $K$ is a free parameter that must be chosen before the model can be applied to data. For the WSBM, we must also choose the tuning parameter $\alpha$ and the exponential family distributions $(T,\eta)$.

In principle, any of several model selection techniques could be used, including minimum description length~\cite{peixoto_2012_parsimonious}, integrated likelihood~\cite{come_model_2013} or Bayes factors~\cite{hofman_bayesian_2008}. Classic complexity-control techniques like the AIC or BIC are known to misestimate $K$ in certain situations~\cite{yan_model_2012}.
Here, we describe an approach for choosing $K$ based on Bayes factors that chooses the value $K$ with largest marginal log-likelihood.

Let $\mathcal{M}_1 = (K_1,\alpha_1,T,\eta)$ and $\mathcal{M}_2 = (K_2,\alpha_2,T,\eta)$ be two competing models, one with $K_{1}$ groups and one with $K_{2}$ groups. 
The Bayes factor between these models is
\begin{equation}
\log B(\mathcal{M}_1,\mathcal{M}_2) = \log\frac{\Pr(A \,|\, \mathcal{M}_1)}{\Pr(A \,|\, \mathcal{M}_2)} \approx {\G}_1 - {\G}_2 \enspace,
\end{equation}
where we approximate the marginal log-likelihood of each model $\Pr( A \,|\, \mathcal{M}_i)$ with our lower bound $\G_i$ Eq.~\eqref{eq:lowerboundG}.
Although Bayes factors assigns a uniform prior on a set of nested models, this approach has a built-in penalty for complex models through the prior distribution. 
In our experiments below, we treat $K,\alpha,T,\eta$ as fixed. This method has produced good results on synthetic data with known planted structure~\cite{aicher_adapting_2013}.

\begin{figure}[tb] 
	\centering
		\subfigure[{Example Network $\sigma^2 = 0.15$}]{
			\raisebox{17pt}{\includegraphics[width = .25\textwidth, clip = true, trim = 0.5in 2.5in 0.5in 3in]{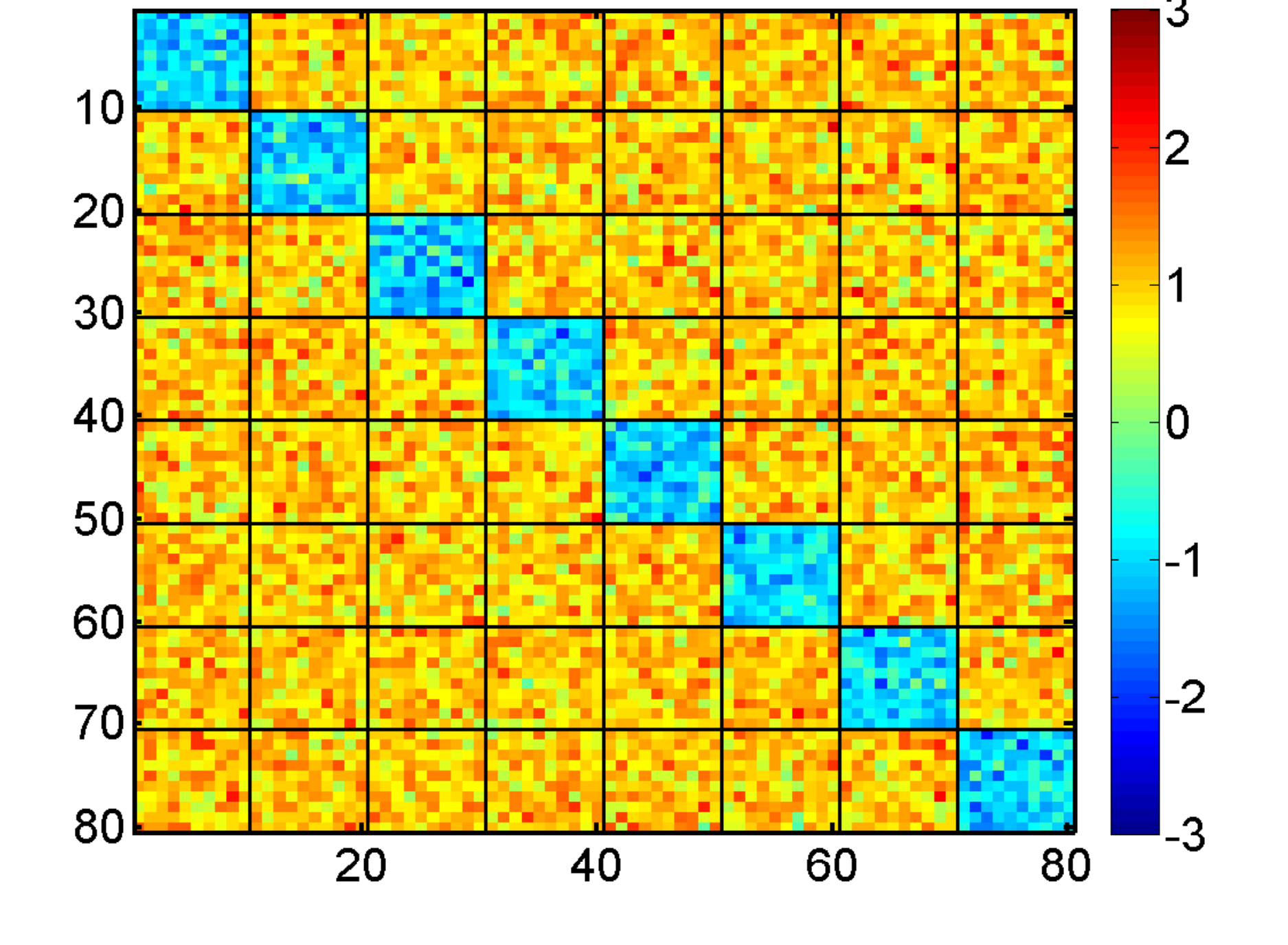}}
			\label{fig:exK8Network}
			}
		\subfigure[{$\G$ vs $K$ Inferred}]{
		\centering
			\includegraphics[width = .39\textwidth, clip = true, trim = 0in 2in 0in 3.85in]{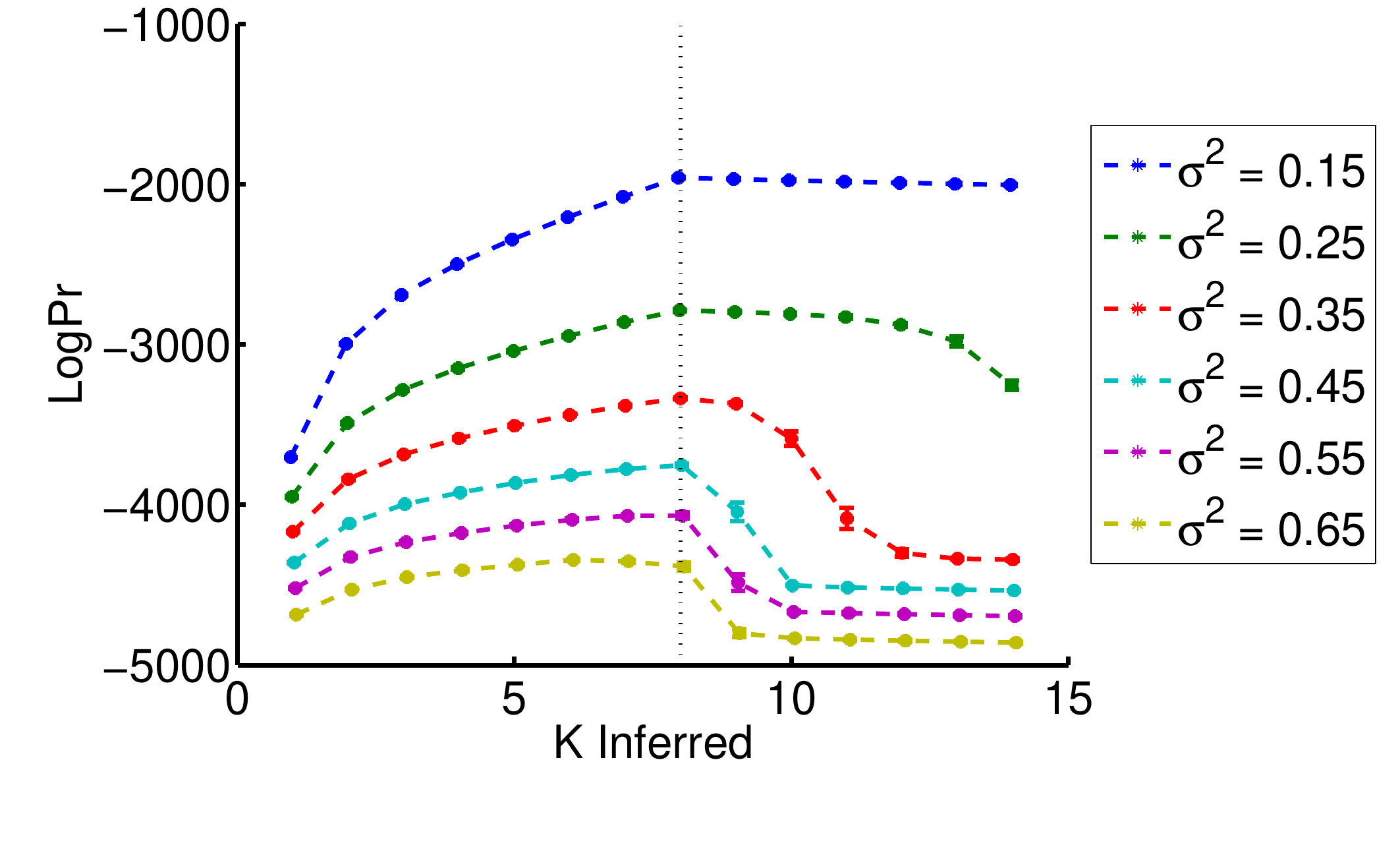}
			\label{fig:K8LogPrvsK}
			}
		\subfigure[{NMI vs $K$ Inferred}]{
		\centering
			\includegraphics[width = .30\textwidth, clip = true, trim = 0in 2in 1.88in 3.85in]{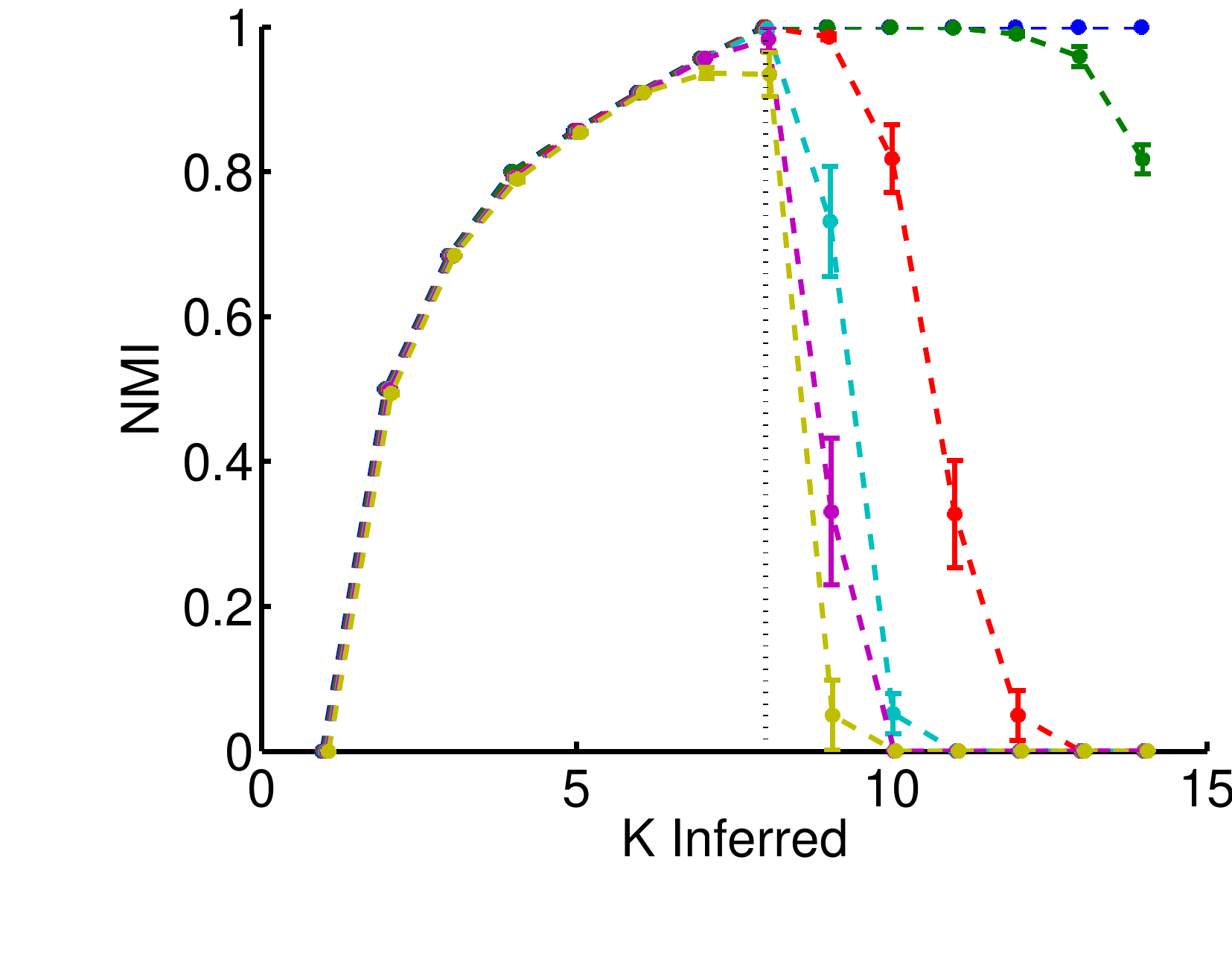}
			\label{fig:K8NMIvsK}
			}
		\caption{
		(a) Example network with $K = 8$ groups and variance $\sigma^2 = 0.15$.
		(b) Approximate marginal log-likelihood $\G$ for each model as a function of $K$.  
		(c) NMI between the fitted model and the true planted structure as a function of $K$.
		Each data series lines in (b,c) corresponds to a different choice of variance $\sigma^{2}$ in edge weights.
		Results are averaged over 20 trials and the error bars are the standard errors.
		}
		\label{fig:K8ModelSelect}
\end{figure}  

We now demonstrate the use and efficacy of Bayes factors in selecting the number of groups $K$ in the WSBM. 
For our demonstration, we choose $K=8$ groups of $10$ vertices each, and consider the method's performance for a variety of edge-weight structures.
Specifically, edge weights within each group are drawn from a normal distribution with mean $-1$ and variance $\sigma^2$, while edge weights between groups are drawn from a normal distribution with mean $1$ and variance $\sigma^2$. 
By varying the variance parameter $\sigma^2$, we vary the difficulty of recovering the true group structure, with a larger variance $\sigma^2$ making inference more difficult by causing the edge weight distributions within and between groups to increasingly overlap.
Figure~\ref{fig:exK8Network} shows an example network drawn from this model, where we choose $\sigma^2 = 0.15$.

To each choice of $\sigma^{2}$, and for a large number of networks drawn from this model, we fit the WSBM using the normal distribution for the edge weights and vary the number of inferred groups $K$ from $1$ to $14$.
Figure~\ref{fig:K8LogPrvsK} shows the approximate marginal log-likelihood $\G$ of each fitted model as $K$ varies, which represents our proportional belief that each choice of $K$ is the correct. 
Similarly, figure~\ref{fig:K8NMIvsK} shows the NMI between each fitted model and the true planted structure, which represents the performance of each choice of $K$. 
Reassuringly, both quantities are maximized at or close to the true value of $K$.
When the within- and between-group edge-weight distributions are relatively well separated, both the marginal log-likelihood $\G$ and NMI are consistently maximized at $K = 8$, indicating that Bayes factors provide a reasonably reliable method for selecting the correct number of groups and thereby recovering the true planted structure in most cases. As the distributions overlap (greater $\sigma^{2}$ here), it becomes more difficult to distinguish groups, and accuracy degrades to some degree, as would be expected.

 
\section{Experimental Evaluation} 
In this section, we evaluate the performance of the WSBM on several real-world networks, in two different ways. First, we consider the question of whether adding edge-weight information necessarily reinforces the latent group structure contained in the edge existences. That is, can the WSBM can find structure distinct from what the SBM would find?
Second, we evaluate the WSBM's performance on two prediction tasks. The first focuses on predicting missing edges (also called ``link prediction''), while the second focuses on predicting missing edge weights. We compare its performance with other block models through cross-validation.

\subsection{Edge weight versus edge existence latent group structure}
\label{sec:nfl}
\begin{figure}[tb!] 
	\centering
		\subfigure[{SBM ($\alpha\! =\! 1$)}]{
			\includegraphics[width = .4\textwidth, clip = true, trim = 0in 2in 0in 2in]{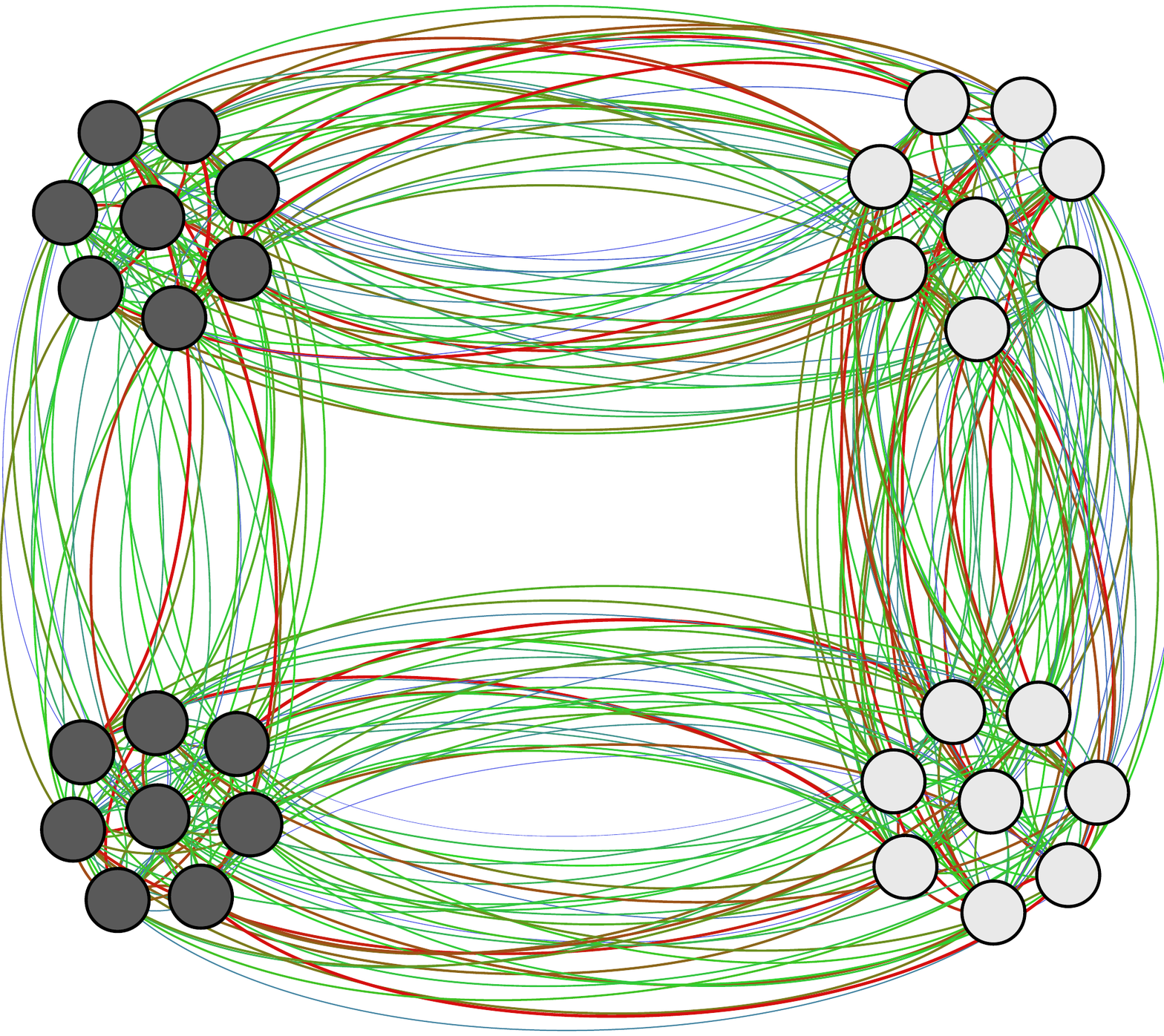}
			\label{fig:NFLSBM}
			}
		\subfigure[{SBM Adjacency Matrix}]{
			\includegraphics[width = .5\textwidth, clip = true, trim = 0in 0in 0in 0in]{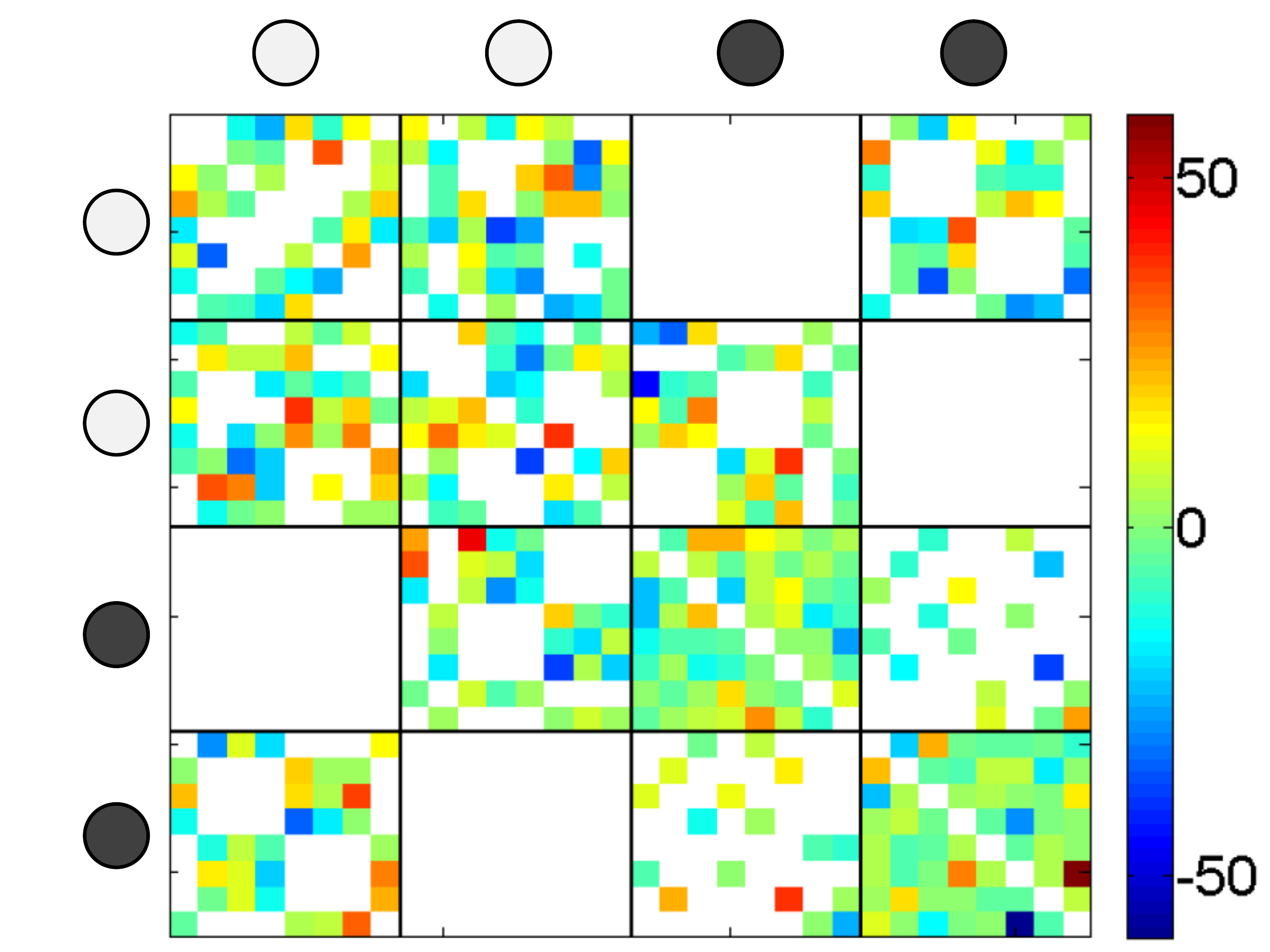}
			\label{fig:NFLSBMadj}
			}
			
		\subfigure[{WSBM ($\alpha\! =\! 0$)}]{
			\includegraphics[width = .4\textwidth, clip = true, trim = 0in 1.5in 0in 2in]{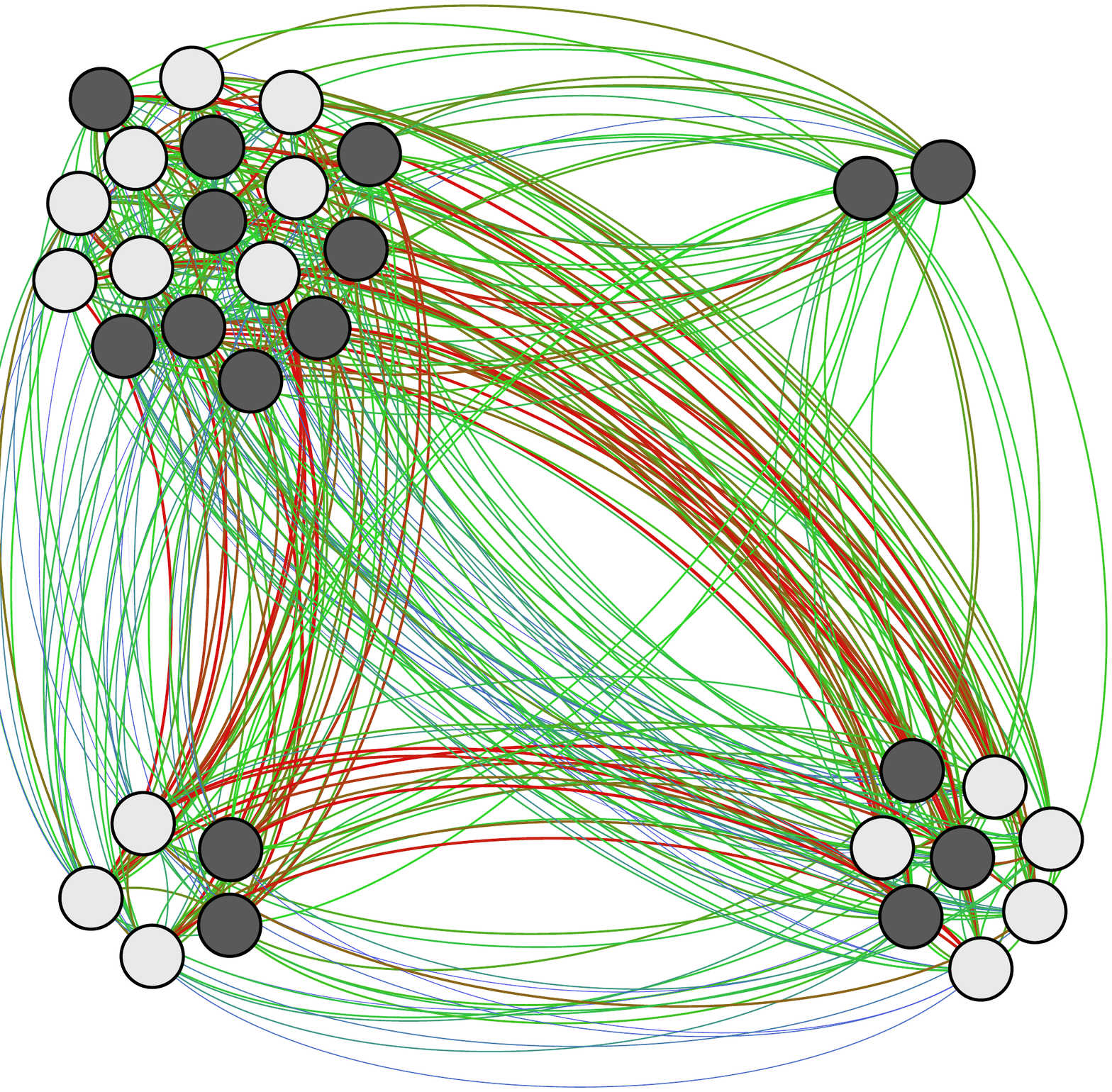}
			\label{fig:NFLWSBM}
			}
		\subfigure[{WSBM Adjacency Matrix}]{
			\includegraphics[width = .5\textwidth, clip = true, trim = 0in 0in 0in 0in]{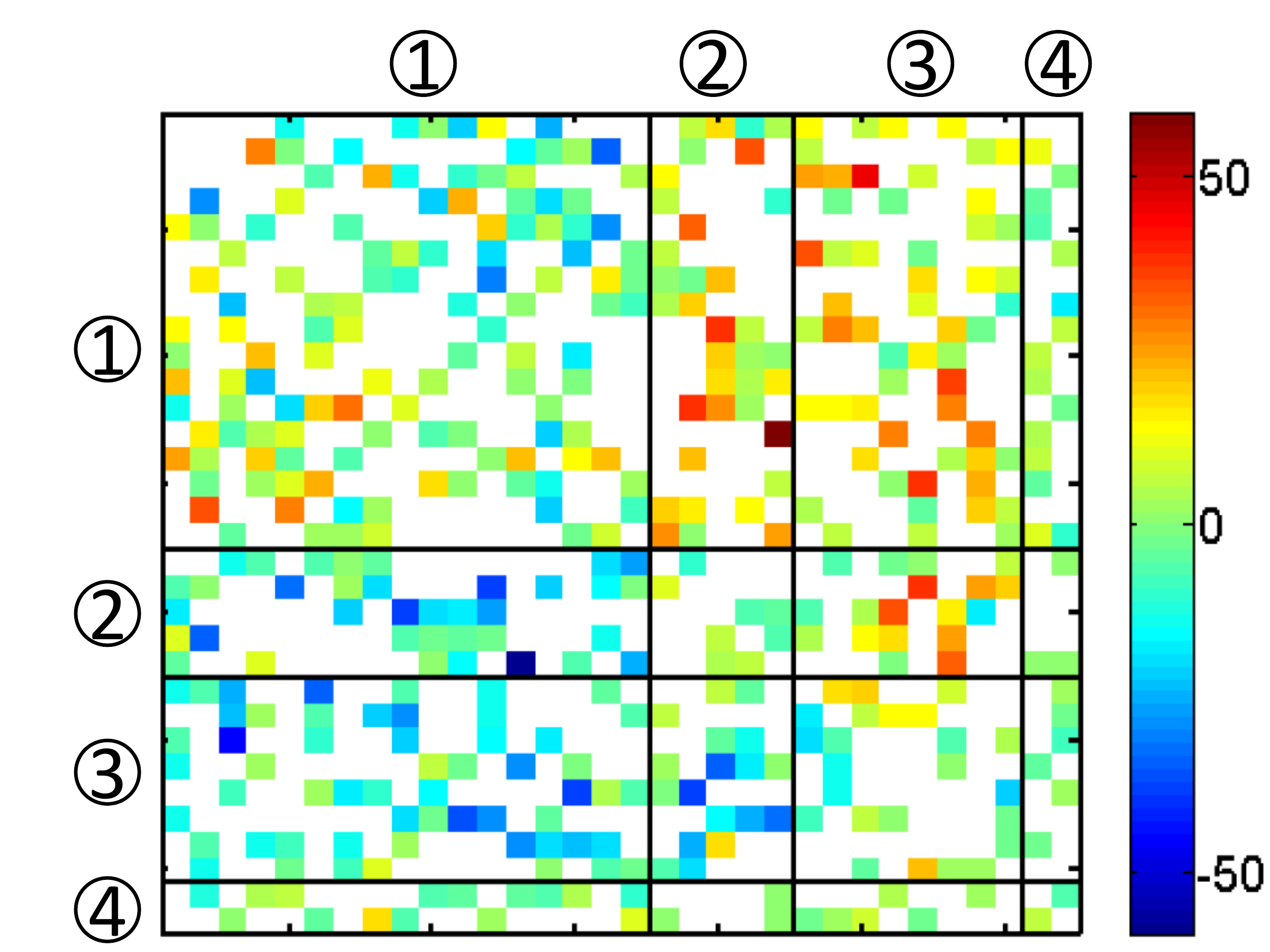}
			\label{fig:NFLWSBMadj}
			}
		\caption{
		NFL-2009 network: black nodes ($\bullet$) are teams in conference 1 (NFC) and white nodes ($\circ$) are teams in conference 2 (AFC). 
		Edges are colored by score differential (red positive, green approximately zero, blue negative).
		(a) Network showing SBM communities. 
		(b) Adjacency matrix, sorted by SBM communities. 
		(c) Network showing WSBM communities. 
		(d) Adjacency matrix, sorted by WSBM communities.
		The SBM ($\alpha\! =\! 1$) groups correspond to NFL conference structure whereas the WSBM ($\alpha\!=\!0$) corresponds to relative skills levels.
		}
		\label{fig:NFL}
\end{figure} 

To probe the question of whether edge weights can contain latent group structures that are distinct from those contained in the edge existences, we consider a simple network derived from the competitions among a set of professional sports teams. In this network, called ``NFL-2009'' hereafter, each vertex represents one of the 32 professional American football teams in the National Football League (NFL). In this network, an edge exists whenever pair of teams played each other in the 2009 season, and each of these edges is assigned a weight equal to the average score difference across games played by that pair~\cite{sport_scoring_2013}. (This definition of edge weight implies the network is skew-symmetric $A_{ij} \!=\! -A_{ji}$.)
These teams are divided equally among two ``conferences'' (called AFC and NFC), and within each conference, teams are assigned to one of 4 divisions, each containing 4 teams. Play among teams, i.e., the existence of an edge, is determined by division memberships, and many teams never play each other during the regular season.

To analyze this network, 
we choose $K\!=\!4$ and fit both the SBM ($\alpha \!=\! 1$) and the ``pure'' WSBM ($\alpha \!=\! 0$) using the normal distribution as a model of edge weights. This choice is reasonable for these data as score differences can be positive or negative and score totals are close to a binomial distribution~\cite{merritt:clauset:2014}.
The $\alpha \!=\! 1$ (SBM) case ignores the weights of edges, while the $\alpha \!=\! 0$ (pure WSBM) case ignores the presence or absence of edges, focusing only on the observed score differences.

Examining the results of both models, we see that the block structure learned by the SBM ($\alpha \!=\! 1$, Figs~\ref{fig:NFL}(a-b)) exactly recovers the
major divisions within each conference, along with the division between conferences, illustrating that division membership fully explains which teams played each other in this season. 
That is, the empty off-diagonal blocks (Fig.~\ref{fig:NFLSBMadj}) reflect the fact that two pairs of two divisions never play each other.

In contrast, the block structure learned by the pure WSBM ($\alpha \!=\! 0$, Figs~\ref{fig:NFL}(c-d)) recovers a global ordering of teams (as in Fig.~\ref{fig:OrderedSBM}) that reflects each team's general skill, so that teams within each block have roughly equal skill. This pattern mixes teams across conference and division lines, and thus disagrees with the block structure recovered by the SBM.
For instance, consider the upper-left group in Fig.~\ref{fig:NFLWSBM}, which generally has positive score differences (wins) in games against teams in either lower group, with a mean lead of 11 points. Similarly, the lower-left group has positive score differences (wins) against teams in the lower-right group. 
The small upper-right group performs equally well against teams of every other group.
Within each group, however, score differences tend toward zero, indicating roughly equal skill.

The fact that the SBM and pure WSBM recover entirely distinct block structures illustrates that adding edge-weight information to the inference step can dramatically alter our conclusions about the latent block structure of a network. That is, adding edge weights does not necessarily reinforce the inferences produced from binary edges alone. The extremal settings of the parameter $\alpha$ in our model allows a practitioner to choose which of these types of latent structure to find, while if a mixed-type conclusion is preferred, an intermediate value of $\alpha$ may be chosen. In the following section, we demonstrate that such a model, which we call the ``balanced'' WSBM, that can learn simultaneously from edge existence and weight information.

\subsection{Predicting edge existence and weight}
To illustrate a more rigorous evaluation of the WSBM, in this section, we consider the problem of predicting missing information when the model is fitted to a partially observed network. In particular, we consider predicting the existence or the weight of some unobserved interaction, a similar task to missing and spurious link prediction~\cite{guimera_missing_2009,clauset_structural_2007}. 

Here, we compare the WSBM to other block models on five real-world networks from various domains. Most of these models are only defined for unweighted networks, and thus some care is required to make them perform under the edge-weight prediction task, which we describe below.
We evaluate performance numerically across multiple trials of cross-validation, training each model on 80\% of the $n^2$ possible edges and testing on the remaining 20\%.

The weighted graphs we consider are the following.
\begin{itemize}[noitemsep,nolistsep]
\item \emph{Airport}. Vertices represent the $n=500$ busiest airports in the United States, and each of the $m=5960$ directed edges is weighted by the number of passengers traveling from one airport to another~\cite{colizza_reactiondiffusion_2007}.
\item \emph{Collaboration}. Vertices represent $n=226$ nations on Earth, and each of the $m=20616$ edges is weighted by a normalized count of academic papers whose author lists include that pair of nations~\cite{pan_world_2012}.
\item \emph{Congress}. Vertices represent the $n=163$ committees in the 102nd United States Congress, and each of the $m=26569$ edges is weighted by the pairwise normalized ``interlock'' value of shared members~\cite{porter_committees_2005}.
\item \emph{Forum}. Vertices represent $n=1899$ users of a student social network at UC Irvine, and each of the $m=20291$ directed edges is weighted by the number of messages sent between users~\cite{opsahl_email_2009}.
\item \emph{College FB}. Vertices represent the $n=1411$ NCAA college football teams, and each of the $m=22168$ edges are weighted by the average point difference across games between a pair of teams~\cite{sport_scoring_2013}.
\end{itemize}
For each of the two prediction tasks and for each network, we evaluate the following models. The ``pure'' WSBM (pWSBM), using only weight information $(\alpha \!=\! 0)$, a ``balanced'' WSBM (bWSBM), using both edge and weight information $(\alpha \!=\! 0.5)$, the ``classic'' SBM, using only edge information $(\alpha \!=\! 1)$, a degree-corrected weighted block model DCWBM, where $(\alpha \!=\! 0.5)$ and the degree-corrected block model (DCBM). 
For the weighted block models, we select the normal distribution to model the edge weights.

In both prediction tasks, we first choose a uniformly random 
$20\%$ of the $n^2$ interactions, which we treat as missing when we fit the model to the network. 
We then fit each model to the observed edges and infer group membership labels for each vertex in the network.  
Finally, we use the posterior mean obtained from variational inference as the predictor for edge existence and edge weight for unobserved interactions between those groups.
For the models that do not naturally model edge weights (SBM, DCBM), we take their partitions and compute the sample mean weight for each of the induced edge bundles in the weighted network and use this value to predict the weight of any missing edge in that bundle.
These estimators correctly correspond to the underlying generative model for edge-prediction in the SBM and DCBM, and are a natural extension for predicting edge-weights for a given block membership.
Under this scheme, each model is made to predict the unobserved interactions for a given network, and we score the accuracy of these predictions using the mean-squared error (MSE).
Evaluating edge-existence prediction could be achieved using alternative criteria such as AUC~\cite{clauset:moore:newman:2008}, which gives similar results.  

Each of these models has a free parameter $K$ that determines the number of parameters that are estimated, which thus controls their overall flexibility. We control this variable model complexity and ensure a fair comparison by fixing all models to have $K \!=\! 4$ latent groups, and we treat all networks as directed. 
Finding the true number of latent groups $K$ for each network is separate worthwhile problem not considered here.  
To compare the results across different data sets, all edge-weights were normalized to fall on the interval $[-1,1]$. Non-negative weights were normalized after applying a logarithmic transform (cases marked with a star $*$ in Tables~\ref{tab:EdgeMSE} and~\ref{tab:WeightMSE}).
 
\begin{table*}[t]
\caption{Average mean-squared error (MSE) on edge prediction in 25 trials.}
\vskip 0.05in
\begin{center}
\small
\begin{tabular}{| l | c | c | c | c | c | c | c |}
\hline 
Network & pWSBM & bWSBM & SBM & DCWBM & DCBM \\
\hline
Airport & 0.0202(1)\hspace{0.5em} & \textbf{0.0156(1)}\hspace{0.5em} & \textbf{0.0158(1)}\hspace{0.5em} & 0.0238(1)\hspace{0.5em} & 0.0238(1) \\
Collaboration\hspace{0.5em} & 0.1446(3)\hspace{0.5em} & 0.1167(3)\hspace{0.5em} & \textbf{0.1138(3)}\hspace{0.5em} & 0.2289(5)\hspace{0.5em} & 0.2454(5) \\
Congress & 0.1765(4)\hspace{0.5em} & \textbf{0.1648(4)}\hspace{0.5em} & \textbf{0.1640(5)}\hspace{0.5em} & 0.2298(9)\hspace{0.5em} & 0.2402(9) \\
Forum & 0.00560(1) & \textbf{0.00535(1)} & \textbf{0.00535(1)} & 0.00565(1) & 0.00565(1)\hspace{-0.5em} \\
College FB & 0.0369(2)\hspace{0.5em} & \textbf{0.0344(1)}\hspace{0.5em} & \textbf{0.0346(1)}\hspace{0.5em} & 0.0387(2)\hspace{0.5em} & 0.0389(2) \\
\hline
\end{tabular}
\end{center}
\label{tab:EdgeMSE}
\caption{Average mean-squared error (MSE) on normalized weight prediction in 25 trials.}
\vskip 0.05in
\begin{center}
\small
\begin{tabular}{| l | c | c | c | c | c | c | c |}
\hline 
Network & \hspace{0.3em}pWSBM\hspace{0.5em} & \hspace{0.45em}bWSBM\hspace{0.45em} & \hspace{0.9em}SBM\hspace{1.3em} & \hspace{0.2em}DCWBM\hspace{0.5em} & \hspace{0.8em}DCBM\hspace{0.2em} \\
\hline
Airport* & \textbf{0.0486(6)} & 0.0543(5) & 0.0632(8) & 0.0746(9) & 0.0918(8) \hspace{-0.5em}\\
Collaboration*\hspace{0em} & \textbf{0.0407(1)} & 0.0462(1) & 0.0497(3) & 0.0500(2) & 0.0849(3) \hspace{-0.5em}\\
Congress* & \textbf{0.0571(4)} & 0.0594(4) & 0.0634(6) & 0.0653(4) & 0.1050(6) \hspace{-0.5em}\\
Forum* & \textbf{0.0726(3)} & 0.0845(3) & 0.0851(4) & 0.0882(4) & 0.0882(4) \hspace{-0.5em}\\
College FB & \textbf{0.0124(1)} & 0.0140(1) & 0.0145(1) & 0.0149(1) & 0.0160(2) \hspace{-0.5em}\\
\hline
\end{tabular}
\end{center}
\label{tab:WeightMSE}
\end{table*}  	
For each model and each network, we ran 25 independent trials with our $80/20$ cross-validation split, as described above, and then compute the average MSE on the particular prediction task.
The results for predicting edge existences are summarized in Table~\ref{tab:EdgeMSE} and the results for predicting edge weights are summarized in Table~\ref{tab:WeightMSE}. 
Bolded values denote the best MSE across all models, and parentheses indicate the uncertainty (standard error) in the last digit.

Notably, in the edge-existence prediction task, the SBM and the balanced WSBM are the most accurate among all models, often by a large margin. The fact that the SBM performs well is perhaps unsurprising, as it is, by design, only sensitive to edge existences in the first place. However, the balanced WSBM is learning from both existence and weight information, and its strong performance indicates that for these networks, learning from edge weights does not necessarily confuse predictions on edge existence. In the edge-weight prediction task, however, the pure WSBM ($\alpha \!=\! 0$) is the most accurate, often by a large margin, as we might expect for a model designed to learn only from edge weight information.

In this experimental framework, none of the degree corrected models performs well.
This is likely caused by the DCBM's and DCWBM's correction for edge propensity in the group membership. By focusing on finding community structure after accounting for edge propensity, the DCBM and DCWSBM have less accurate predictions in predicting edge existence and edge weight. 
It is worth pointing out, however, that prediction is not the only measure of utility for community detection techniques, and degree-corrected models often perform better than non-corrected models at recovering meaningful latent group structures in practical situations. We thus expect the degree-corrected WSBM will be most useful in situations where the goal is the recovery of scientifically meaningful group structures, rather than edge existence or weight prediction.

In general, the SBM performs well on edge prediction but poorly on weight prediction, while the pure WSBM performs poorly on edge prediction but well on weight prediction. 
This pattern is precisely as we might expect, as the SBM only considers existence information, while the pure WSBM only considers weights.

What is surprising, however, is the good performance on both tasks by the balanced WSBM ($\alpha \!=\! 0.5$), which is as good or nearly as good as SBM in edge prediction, but substantially better than the SBM in weight prediction. 
This demonstrates that the balanced WSBM is a more powerful model than the SBM: it performs as well as the SBM on SBM-like tasks and better on edge weight tasks. 
In these examples, incorporating edge weight information into the SBM framework does not detract the WSBM performance in edge prediction. 
In fact, this good general performance is possible because the balanced WSBM learns from both edge existence and edge weight information.


\section{Discussion} 

In the analysis of networks, the inference of latent community structure is a common task that facilitates subsequent analysis, e.g., by dividing a large heterogeneous network into a set of smaller, more homogeneous subgraphs, and can reveal important insights into its basic organizational patterns. When edges are annotated with weights, this extra information is often discarded, e.g., by applying a single universal threshold to all weights.
The weighted stochastic block model (WSBM) we described here is a natural generalization of the popular stochastic block model (SBM) to edge-weighted sparse networks. Crucially, the WSBM provides a statistically principled solution to the community detection problem in edge-weighted networks, and removes the need to apply any thresholds before analysis. Thus, this model preserves the maximal amount of information in such networks for characterizing their large-scale structure.

The WSBM's general form, given in Eq.~\eqref{eq:general}, is parametrized by a mixing parameter $\alpha$, which allows it to learn simultaneously from both the existence (presence or absence) of edges and their associated weights. In our tests with real-world networks, the WSBM yields excellent results on both edge existence and weight prediction tasks. Additionally, the balanced model ($\alpha\!=\!0.5$) performed as well or nearly as well as the best alternative block model, suggesting it may work well as a general model for novel applications where it is not known whether edge existences or edge weights are more informative.

In many applications, the inferred group structure will be of primary interest. 
For these cases, it is important to note that the groups identified by the WSBM can be distinct from those identified by examining only an unweighted version of the same network. 
Both forms of latent structure may be interesting and are likely to shed different light on the underlying organization of the network. 
It remains an open question to determine the types of networks for which weight information contains distinct partition structure from edge existences, although we have shown at least one example of such a network in section~\ref{sec:nfl}.


The variational algorithm described here provides an efficient method for fitting the WSBM to an empirical network. 
It scalability is relatively good by modern standards, and thus should be applicable to networks of millions of vertices or more. 
Alternative algorithms such as those based on Markov chain Monte Carlo for unweighted networks are possible~\cite{peixoto_2013_hierarchcial,clauset_structural_2007}; however, each must contend with several technical problems presented by edge weight distributions, e.g., the degeneracies in the likelihood function produced by edge-bundles whose weights have zero variance. 


Finally, there are several natural extensions of the WSBM, including mixed memberships~\cite{airoldi_mixed_2008}, bipartite forms~\cite{larremore_efficiently_2014}, dynamic networks~\cite{peel_detecting_2014}, different distributions for different edge bundles, and the handling of more complex forms of auxiliary information, e.g., on the vertices or edges.
An important and open theoretical question presented by this model is whether utilizing weight information modifies the fundamental detectability of latent group structure, which exhibits a phase transition in the classic SBM~\cite{decelle_phasetransition_2011}. 
We look forward to these and other extensions.

\section*{Funding}
This work was supported by the U.S.\ Air Force Office of Scientific Research and the Defense Advanced Research Projects Agency [grant number FA9550-12-1-0432]. 

\section*{Acknowledgments}
We thank Dan Larremore, Leto Peel, and Nora Connor for helpful conversations and suggestions.

Certain data included herein are derived from the Science Citation
Index Expanded, Social Science Citation Index and Arts \& Humanities
Citation Index, prepared by Thomson Reuters, Philadelphia, Pennsylvania,
USA, Copyright Thomson Reuters, 2011

\newpage
\appendix
\section{Code Availability}
\label{app:code}
A working implementation of the WSBM inference code, written by the authors, may be found at \url{http://tuvalu.santafe.edu/%7Eaaronc/wsbm/}. 

This code implements the efficient algorithms discussed in Appendix~\ref{app:sparseweighted}.

\section{Exponential Families}
\label{app:expfamily}
Let $\mathcal{X}$ be a fixed domain and $\Theta$ be set of parameters.
An exponential family is a collection of parametric distributions $\F$ that can be written in the form 
\begin{equation*}
\F = \{ f(x \,|\, \theta) = h(x)\exp\left(T(x) \cdot \eta(\theta)\right) \text{ for } x \in \mathcal{X} \, | \, \theta \in \Theta \} \enspace,
\end{equation*}
where $h,T,\eta$ are fixed functions.
The map $T$ is the sufficient statistic function and the map $\eta(\theta)$ are the natural parameters. 
Note that $T$ and $\eta$ can be vectors.  
The function $h(x)$ distinguishes different probability distributions, but appears as an additive constant in the log-likelihood function, which can thus be ignored. Thus, only the pair $(T,\eta)$ directly impacts the likelihood function. 

Examples of exponential families include the normal, exponential, gamma, log-normal, Pareto, binomial, multinomial, Poisson, and beta distributions.
Examples of distributions that are not exponential families are the Uniform distribution and certain mixture distributions.

A common representation of an exponential family sometimes includes the log-partition function $A(\theta)$ written as
\begin{equation*}
f(x \,|\, \theta) = h(x) \exp\left(\tilde{T}(x) \cdot \tilde{\eta}(\theta)-A(\theta)\right) \enspace.
\end{equation*}
To keep notation compact we absorb $-1 \cdot A$ into $T \cdot \eta$.

A convenient property of exponential families is that they have easily written conjugate priors. 
For our exponential family the standard class of conjugate priors $\pi$ are
\begin{equation*}  
\pi(\theta) = \frac{1}{Z(\tau)}\exp\left(\tau \cdot \eta(\theta)\right) \enspace, 
\end{equation*}
where $\tau$ are the (hyper-)parameters of the prior and can be thought of as pseudo-observations of $T$.
The function $Z$ is the normalizing constant, defined as
\begin{equation*}
Z(\tau) = \int_\Theta \exp\left(\tau \cdot \eta(\theta)\right) \ d\theta \enspace.
\end{equation*}
Finally, it can be shown that the expected value of $\eta(\theta)$ under $\pi(\cdot \,|\, \tau)$ is
\begin{equation*}
\left\langle\eta(\theta)\right\rangle = \frac{\partial \log Z(\tau)}{\partial \tau} \enspace.
\end{equation*}
Further details on exponential families can be found in Refs.~\cite{ohagan_kendalls_2009,robert_bayesian_2007} and for appropriate prior distributions in Ref.~\cite{berger_development_1992}. 

\section{Belief Propagation Derivation}
\label{app:bp}
The main difference between a loopy belief propagation (hereafter simply BP) algorithm and the variational Bayes algorithm described in the main text lays in how we update the group membership parameters $\mu$~\cite{yedidia_understanding_2003}. 
The BP approach gives a more accurate approximation of the true posterior of $z$ and has been shown to produce good results in the classic SBM case~\cite{decelle_phasetransition_2011}. 

In variational Bayes, we used a mean-field approximation to the posterior distribution $\pi^*$:
\begin{equation*}
\pi^*(z) \approx q(z) = \prod_i \mu_i(z_i) \enspace,
\end{equation*} 
where each vertex label is assumed to be independently distributed according to $q_i$ (a categorical or multinomial random variable).

In BP, we use pairwise approximations to the posterior distribution. Ideally, this approximation would have the form
\begin{equation*}
\pi^*(z) \approx q(z) \propto \prod_{ij} \mu_{ij}(z_i,z_j) \enspace ,
\end{equation*} 
where $\mu_{ij}(\cdot,\cdot)$ are joint probabilities. 
However, this typically is not achievable because normalizing the product of distributions over all edge pairs is non-trivial (each vertex of degree $k_i$ appears $k_i$ times). 
Luckily, in the case of trees, it is possible to normalize $q$ to a probability distribution by accounting for this repetition, that is, 
\begin{equation*}
q(z) = \frac{\prod_{ij \in E} \mu_{ij}(z_i,z_j)}{\prod_i \mu_i(z_i)^{k_i-1}} \enspace,
\end{equation*} 
where $\mu_i$ is the marginal of $\mu_{ij}$, $k_i$ is the degree of vertex $i$, and $E$ is the set of observed edges.
But, the factor graph of the WSBM is not a tree, so this form is not necessarily exact.

Here, we take a loopy BP approach and assume the structure of pairwise terms $ij \in E$ is in fact locally tree-like, and then apply the BP update equations. 
The assumption for locally tree-like structure makes this algorithm a poor choice on dense networks (when we observe $O(n^2)$ interactions), but is both acceptable and effective for sparse networks.

Under this formulation, our goal is to maximize the variational approximation to the likelihood of the data $\G$, so that the KL divergence between $q$ and $\pi^*$ is minimized. 
Recall from Eq.~\eqref{eq:Gp} the objective function $\G$ consists of two parts
\begin{equation*}
\G = \E_q\log \Pr(A \,|\, z,\theta) + \E_q \log \left(\pi / q\right)  \enspace,
\end{equation*}
a likelihood term and a prior regularizer term.

The likelihood term is
\begin{align*}
\E_q\log \Pr(A\,|\,z,\theta) &\propto \sum_r \left(\sum_{ij} T(A_{ij}) \E_q(z_i z_j) + \priortau_r \right) \cdot \left\langle\eta\right\rangle_r \approx \sum_r \left( \left\langle T \right\rangle_r + \priortau_r \right) \cdot \left\langle\eta\right\rangle_r \enspace,   
\end{align*}
where
\begin{align*}
  \left\langle T \right\rangle_r &= \sum_{ij} \sum_{(z,z') = r} \mu_{ij}(z,z') \, T(A_{ij}) \\
  \left\langle \eta \right\rangle_r &= \left.\frac{\partial \log Z(\tau)}{\partial \tau}\right|_{\tau = \tau_r}  \enspace, 
\end{align*}
and where we approximate $\E_q(z_i z_j) \approx q_{ij}(z_i,z_j)$ and $\priortau_r, \priormu$ are parameters for the prior.

The regularizer term consists of two parts
\begin{equation*}
\E_q \log\left(\pi / q\right) = \E_q\left(\log\pi\right) - \E_q\left(\log q \right) \enspace. 
\end{equation*}
The second term is requires us to sum over $q(z)$ which is combinatorically difficult to calculate, so we use the Bethe approximation
\begin{equation*}
-\E_q\left(\log q\right) \approx -\sum_{ij \in E}\sum_{z,z'} \mu_{ij}(z,z')\log\mu_{ij}(z,z')+ \sum_{i,z} (k_i-1)\mu_i(z)\log\mu_i(z) + \sum_r -\tau_r \cdot \left\langle\eta\right\rangle_r + \log Z(\tau_r) \enspace .
\end{equation*}
Combining these parts, the objective function may be written as
\begin{align*}
\G =& \sum_r \left(\left\langle T \right\rangle_r +\priortau_r-\tau_r\right) \cdot \left\langle\eta\right\rangle_r + \sum_r \log \frac{Z(\tau_r)}{Z(\priortau_r)}\\
 & + \sum_{i,z} (k_i-1)\mu_i(z)\log\frac{\mu_i(z)}{\priormu_i(z)} - \sum_{ij \in E} \sum_{z,z'} \mu_{ij}(z,z')\log\frac{\mu_{ij}(z,z')}{\priormu_i(z) \priormu_j(z')}  \enspace.
\end{align*}

To enforce the marginalization and normalization restrictions on $q(z)$, we introduce Lagrange multipliers, yielding
\begin{equation*}
\G' = \G+ \sum_i \lambda_i \left(\sum_i \mu_i - 1\right) + \sum_{ij \in E} \left( \sum_z \lambda_{ij,z} \left(\mu_i(z)-\sum_{z'} \mu_{ij}(z,z')\right)+\sum_{z'} \lambda'_{ij,z'} \left(\mu_j(z')-\sum_{z} \mu_{ij}(z,z')\right) \right) \enspace .
\end{equation*}
Note that $\lambda_i$ enforces normalization of $\mu_i$, $\lambda_{ij,z}$ enforces marginalization over $i$, and $\lambda'_{ij,z'}$ enforces marginalization over $j$. 
We maximize $\G'$ by setting its derivatives with respect to the parameters of $q$ equal to $0$

For the edge parameters $\theta$, we differentiate with respect to $\tau_r$
\begin{align*}
\frac{\partial \G'}{\partial \tau_r} &= \left(\left\langle T \right\rangle_r + \priortau_r - \tau_r\right) \frac{\partial \left\langle \eta \right\rangle_r}{\partial \tau_r} - \left\langle \eta \right\rangle_r + \left.\frac{\partial \log Z(\tau)}{\partial \tau}\right|_{\tau = \tau_r} \\
 &\propto \left\langle T \right\rangle_r + \priortau_r - \tau_r \enspace.
\end{align*}
This is the same expression as for the variational Bayes solution, since we only modified $q(z)$.   
The update equations for $\tau$ remain
$\tau_r = \priortau_r+ \left\langle T \right\rangle_r$. 

For the vertex labels $z$, we will differentiate with respect to $\mu_i(z)$ and $\mu_{ij}(z,z')$ and solve this system of equations using a message passing method, which is standard in BP. 
The derivatives are
\begin{equation*}
\frac{\partial \G'}{\partial \mu_i(z)} = (k_i-1)\left(\log\mu_i(z) - \log \priormu_i(z) + 1\right) + \lambda_i + \sum_{j : ij \in E} \lambda_{ij,z} = 0 \enspace ,
\end{equation*}
and
\begin{equation*}
\frac{\partial \G'}{\partial \mu_{ij}(z,z')} = T(A_{ij}) \cdot \left\langle\eta\right\rangle_{z,z'}-\log \mu_{ij}(z,z')+\log \priormu_i(z)+\log \priormu_j(z') - 1 - \lambda_{ij,z}-\lambda'_{ij,z'} = 0 \enspace.
\end{equation*}
Solving for $\mu_i(z)$ and $\mu_{ij}(z,z')$ we obtain
\begin{align*}
\mu_i(z) &\propto \priormu_i(z) \prod_{j: ij \in E}e^{-\lambda_{ij,z}/(k_i-1)}\\
\mu_{ij}(z,z') &\propto \priormu_i(z)\priormu_j(z')\exp\left(T(A_{ij}) \cdot \left\langle\eta\right\rangle_{z,z'}\right) e^{-\lambda_{ij,z}}e^{-\lambda'_{ij,z'}}  \enspace .
\end{align*}
For notational convenience, let 
\begin{equation*}
\evid_{ij}(z,z') = \exp\left(T(A_{ij}) \cdot \left\langle\eta\right\rangle_{z,z'}\right) \enspace .
\end{equation*}
Since $\sum_{z'} \mu_{ij}(z,z') = \mu_i(z)$, we have
\begin{equation*}
\mu_i(z) \propto \priormu_i(z) \sum_{z'} \priormu_j(z') \evid_{ij}(z,z') e^{-\lambda_{ij,z}}e^{-\lambda'_{ij,z'}} \enspace .
\end{equation*}
%
Setting our two equations for $\mu_i(z)$ are equal, we obtain
\begin{align}
\priormu_i(z) \prod_{j': ij' \in E} e^{-\lambda_{ij',z}/(k_i-1)} \propto \priormu_i(z) \sum_{z'} \priormu_j(z') \evid_{ij}(z,z') e^{-\lambda_{ij,z}}e^{-\lambda'_{ij,z'}} \nonumber \\
\prod_{j': ij' \in E} e^{-\lambda_{ij',z}/(k_i-1)} \propto \sum_{z'} \priormu_j(z') \evid_{ij}(z,z') e^{-\lambda_{ij,z}}e^{-\lambda'_{ij,z'}}  \enspace .
\label{eq:mu0}
\tag{*}
\end{align}
Let $\mes_{i\rightarrow j}(z_j)$ denote the message from vertex $i$ to vertex $j$ and set
\begin{equation*}
e^{-\lambda_{ij,z}} = \prod_{k: ik \in E, k \neq j} \mes_{k\rightarrow i}(z) 
\end{equation*}\begin{equation*}
e^{-\lambda'_{ij,z'}} = \prod_{k: j,k \in E, k \neq i} \mes_{k\rightarrow j}(z') \enspace .
\end{equation*}
Plugging in our definition of $\mes$, we obtain
\begin{equation*}
\prod_{j: ij \in E} e^{-\lambda_{ij,z}/(k_i-1)} = \prod_{j: ij \in E} \prod_{k: ik \in E, k \neq j} \mes_{k\rightarrow i}(z)^{1/(k_i-1)} = \prod_{ij \in E} \mes_{j \rightarrow i}(z) \enspace .
\end{equation*}
And, using Eq.~\eqref{eq:mu0}, we obtain the following recusive definition for $\mes$
\begin{align*}
\prod_{ij \in E} \mes_{j \rightarrow i}(z) &\propto \sum_{z'}  \priormu_j(z') \left(\sum_{z'} \evid_{ij}(z,z')\right) \prod_{k: ik \in E, k \neq j} \mes_{k\rightarrow i}(z)\prod_{k: jk \in E, k \neq i} \mes_{k\rightarrow j}(z') \\
\mes_{j \rightarrow i}(z) &\propto \sum_{z'}  \priormu_j(z') \evid_{ij}(z,z') \prod_{k: k,j \in E, k \neq i} \mes_{k\rightarrow j}(z') \enspace.
\end{align*}
Finally, our update equations for $\mu$ become
\begin{align*}
\mu_i(z) &\propto \priormu_i(z) \prod_{ij \in E} \mes_{j \rightarrow i}(z)\\
\mu_{ij}(z,z') &\propto \priormu_i(z)\priormu_j(z')\evid_{ij}(z,z')\prod_{k: ik \in E, k \neq j} \mes_{k\rightarrow i}(z) \prod_{l: l,j \in E, l \neq i} \mes_{l\rightarrow j}(z') \enspace.
\end{align*}

If $m = |E|$ is the number of observed edges/interactions, then the BP algorithm requires $O(m)$ messages to be passed and therefore each iteration has an $O((m+n)K^2)$ running time (updating the messages $\mes$ and then the group membership parameters $\mu$).

It will be convenient to use the following equivalent messages $\mess$ used by~\cite{zhang_comparative_2012,yan_model_2012,decelle_phasetransition_2011} in our BP algorithm
\begin{equation*}
\mess_{i\rightarrow j}(z') = \priormu_j(z') \prod_{k: k,j \in E, k\neq i} \mes_{k,j}(z') \enspace.
\end{equation*}
Note that from our old message $\mes$ update equations, we obtain
\begin{equation*}
\mes_{i\rightarrow j}(z') = \sum_{z} \evid_{ij}(z,z') \mess_{j \rightarrow i}(z) \enspace.
\end{equation*}
Putting these two equations together, our new update equations using $\mess$ for our messages become
\begin{align*}
\mess_{i \rightarrow j}(z') = \priormu_j(z') \prod_{k: kj \in E, k\neq i} \sum_{z} \evid_{k,j}(z,z') \mess_{j \rightarrow k}(z) \\
\mu_i(z) \propto \priormu_i(z) \prod_{ij \in E}\sum_{z'} \evid_{ij}(z,z') \mess_{i \rightarrow j}(z') \\
\mu_{ij}(z,z') \propto \evid_{ij}(z,z') \mess_{j \rightarrow i}(z) \mess_{i \rightarrow j}(z') \enspace.
\end{align*} 
Algorithm \ref{alg:BP} gives pseudocode for the full loopy BP algorithm. 

\begin{algorithm}[h]
   \caption{Loopy BP for sparse networks}
   \label{alg:BP}
\begin{algorithmic}
   \STATE {\bfseries Input:} Data $E$, Model $\M$
   \STATE Initialize $\mu$
   \REPEAT
   \FORALL{$r = 1,\ldots, K^2$}
   \STATE Set $\left\langle T \right\rangle_r := \sum\limits_{ij} \sum\limits_{(z_i,z_j) = r} \mu_i(z_i) \mu_j(z_j) T(A_{ij})$
   \STATE Set $\tau_r := \tau_0 + \left\langle T \right\rangle_r$
   \STATE Set $\left\langle \eta \right\rangle_r := \left. \frac{\partial}{\partial \tau} \log Z(\tau) \right|_{\tau=\tau_r}$ 
   \ENDFOR
   \STATE Calculate $\evid_{ij}$ for all $(ij)$ in $E$
	 \STATE Set $\evid_{ij}(k,k') = \exp(T(A_{ij}) \cdot \left\langle \eta \right\rangle_{k,k'}+T(A_{ji}) \cdot \left\langle \eta \right\rangle_{k',k})$ for all $k,k'$
   \REPEAT
   \FORALL{$(ij)$ in $E$}
   \STATE Set $\mess_{j \rightarrow i}(z_i) \propto \mu_0(z_i) \prod\limits_{k \neq i, kj \in E} \sum\limits_{z_k} \mess_{i \rightarrow k}(z_k) \evid_{ik}(z_i,z_k)$  
   \ENDFOR
   \UNTIL{$\mess$ converge}
	 \FORALL{$i = 1,\ldots,n$}
   \STATE Set $\mu_i(z_i) \propto \mu_0(z_i) \prod\limits_{ij \in E} \sum\limits_{z_j} \mess_{i \rightarrow j}(z_j)\evid_{ij}(z_i,z_j)$
	 \ENDFOR
   \UNTIL{$\mu,\tau$ converge }
   \RETURN $\mu,\tau$
\end{algorithmic}
\end{algorithm}


\section{Modifications for Sparse Weighted Graphs}
\label{app:sparseweighted}

We now consider modifications to our variational Bayes algorithm (Algorithm \ref{alg:VB}) and our BP algorithm (Algorithm \ref{alg:BP}) for the case of sparse weighted graphs discussed in section~\ref{sec:swg}.  

Recall that for a network of $n$ nodes we can partition the $n^2$ interaction into 3 disjoint edge lists $W,N,M$, where $W$ is a list of \emph{weighted edges}, $N$ is a list of \emph{non-edges}, and $M$ is a list of \emph{missing edges} or \emph{unobserved} edges. 
We define the union $E = W \cup N$ as the list of \emph{observed} edges.
Let $m_W = | W |$ be the number of weighted edges, $m_E = | E |$ be the number of observed edges, and $m_M = |M|$ be the number of missing edges. 
Note that $m_E+m_M = |E| + |M| = |A| = n^2$.

Both algorithms we presented require $O(|E| K^2)$ time when updating $\mu$.
If the number of \emph{observed} edges is sparse ($|E| = O(n)$), then no changes are required. 
However it may be the case that the number of weighted edges is sparse ($|W| = O(n)$), while the number of non-edges is dense ($|N| = O(n^2)$). 
In this case, if we assume the number of missing edges is also sparse ($|M| = O(n)$), then we can modify Algorithms \ref{alg:VB} and \ref{alg:BP}, so that running time is once again $O(n K^2)$. The key idea is to exploit the structure of our edge-existence distribution.

First we introduce some notation, then we consider the edge bundle $\tau$ updates, and finally we introduce modifications to the group membership $\mu$ updates.

\paragraph{Notation.}
There are two types of degrees: the degree with respect to weighted edges and degree with respect to observed edges. 
Let $d_W^{-}(i)$ be the in-degree of vertex $i$ with respect to weighted edges. 
Let $d_W^{+}(i)$ be the out-degree of vertex $i$ with respect to weighted edges. 
Let $d_E^{-}(i)$ be the in-degree of vertex $i$ with respect to observed edges. 
Let $d_E^{+}(i)$ be the out-degree of vertex $i$ with respect to observed edges. 

Let our exponential family edge-weight distribution $f_w$ under parameter $\theta_w$ take form
\begin{equation*}
f_w(x \,|\, \theta_w) = h_w(x) \exp\left(T_w(x) \cdot \eta_w(\theta_w)\right)  \enspace,
\end{equation*}
where $h_w,T_w,\eta_w$ are fixed functions.

Let our exponential family edge-existence distribution $f_e$ under parameter $\theta_e$ take the form
\begin{equation*}
f_e(x \,|\, \theta_e) = h_e(x) \exp\left(T_e(x) \cdot \eta_e(\theta_e)\right)  \enspace,
\end{equation*}
where $h_e,T_e,\eta_e$ are fixed functions.

Let $R: K \times K \rightarrow r $ be the mapping between the groups and edge-bundles.

\subsection{Update for $\tau$ (edge distribution)}
The edge bundle updates consist of two steps: (i) calculating the expected sufficient statistic $\left\langle T \right\rangle$ for each edge bundle, and (ii) updating $\tau$ for each edge bundle.

\paragraph{Weighted $\tau_w$.}
For the weighted distribution, the expected sufficient statistic $\left\langle T_w \right\rangle_r$ for all edge bundles $r$ can be calculated using Eq.~\eqref{eq:tauw} for all pairs of groups $(z,z')$, as
\begin{equation}
\label{eq:tauw}
\left\langle T_w \right\rangle_{R(z,z')} +\!\!= \sum_{ij \in W} T_w(A_{ij})\mu_i(z) \mu_j(z') \enspace.
\end{equation}
Since the running time for each pair is dominated by the summation over the set $W$, each iteration over Eq.~\eqref{eq:tauw} takes $O(n+m_W)$
in $O(K^2 (n+m_W))$ time.

\paragraph{Edge existence $\tau_e$.}
To update $T_e$ we note that the sufficient statistic value for a non-edge is typically zero except for the last dimension that takes the value $1$ for observed edges. Knowing that this value is $1$ for all edges lets us calculate $T_e$ without needing to sum over $W \cup N$.
 
Therefore we update $T_e$ using Eq.~\eqref{eq:tauw} for all but the last dimensions of $T_e$. 
For the last dimension we update $T_e$ with
\begin{equation}
\label{eq:taue2}
\left\langle T_e \right\rangle_{R(z,z')} +\!\!= \sum_{ij} \mu_i(z) \mu_j(z') - \sum_{ij \in M} \mu_i(z) \mu_j(z') = \left(\sum_{i} \mu_i(z)\right) \left(\sum_{j} \mu_j(z')\right) - \sum_{ij \in M} \mu_i(z) \mu_j(z')  \enspace,
\end{equation}
which takes $O(K^2 (n+m_M))$ time.

\paragraph{Degree-corrected edge existence $\tau_e$.}
For the degree corrected block model, recall that edge existence distribution is modified slightly by replacing the $T_e(A_{ij}) = 1$ in the last dimension of $T_e$ with the product of $i,j$'s in and out degrees, $T_e(A_{ij}) = d_W^{+}(i) d_W^{-}(j)$. This changes equation \eqref{eq:taue2} by replacing $\mu_i(z)$ with $d_W^{+}(i)\mu_i(z)$ and $\mu_j(z')$ with $d_W^{-}(j)\mu_j(z')$. This gives us
\begin{equation}
\label{eq:taudc2}
\left\langle T_e \right\rangle_{R(z,z')} +\!\!= \left(\sum_{i} d_W^{+}(i)\mu_i(z)\right) \left(\sum_{j} d_W^{-}(j)\mu_j(z')\right) - \sum_{ij \in M} d_W^{+}(i)\mu_i(z) d_W^{-}(j)\mu_j(z')  \enspace.
\end{equation}
The running time remains the same as in the edge existence case.

\subsection{Update for $\mu$ (vertex labels)}

\paragraph{Variational Bayes Algorithm.}
The update for the vertex labels under the variational Bayes algorithm, is to (i) calculate $\frac{\partial \left\langle T \right\rangle_r}{\partial \mu_i(z)}$ and (ii) update $\mu_i$ using
\begin{equation*}
\mu_i(z) \propto \exp \left( \sum_r \frac{\partial \left\langle T \right\rangle_r}{\partial \mu_i(z)} \cdot \left\langle \eta \right\rangle_r \right) \enspace.
\end{equation*}

The rate limiting step is in calculating $\frac{\partial \left\langle T \right\rangle_r}{\partial \mu_i(z)}$. 

For the weighted sufficient statistics $T_w$, we calculate for all pairs $(z,z')$ and for each vertex $i$
\begin{equation}
\label{eq:dtauw}
\frac{\partial \left\langle T_w \right\rangle_{R(z,z')}}{\partial \mu_i(z)} \,+\!\!= \sum_{j \in \partial i^+_W} T_w(A_{ij}) \mu_j(z') \enspace \enspace, \enspace \frac{\partial \left\langle T_w \right\rangle_{R(z',z)}}{\partial \mu_i(z)} \,+\!\!= \sum_{j \in \partial i^-_W} \mu_j(z') T_w(A_{ji}) \enspace,
\end{equation}
where $\partial i^{+}_W$ is the neighborhood formed by the outgoing weighted edges of vetex $i$. 
Since the sum in Eq.~\eqref{eq:dtauw} is over $d_W^{+}(i)$ terms, the running time is $O(K^2\sum_i d_W^{+}(i)) = O(K^2(n+m_W))$. 

Similar to how we updated $\tau_e$, in the edge-existence case we update $\frac{\partial \left\langle T \right\rangle_r}{\partial \mu_i(z)}$ by calculating the entire sum and subtracting away the missing edges. 
Again, we exploit the fact that the last dimension of $T_e$ is $1$ for observed edges, and
\begin{equation}
\label{eq:dtaue2}
\frac{\partial \left\langle T_e \right\rangle_{R(z,z')}}{\partial \mu_i(z)} \,+\!\!= \left(\sum_j \mu_j(z')\right) -\sum_{j \in \partial i^+_M} T_e(A_{ij}) \mu_j(z') , \enspace.
\end{equation}
Calculating Eq.~\eqref{eq:dtaue2} for all vertices has a total $O(K^2(n+m_M))$ running time if we pre-calculate $\sum_j \mu_j(z')$. 

For the degree corrected block model we replace $\mu_j$ with $d^{-}_W(j)\mu_j(z')$ and use Eq.~\eqref{eq:dtaue2}.

\paragraph{Loopy BP Algorithm.}
The update for the vertex labels under the BP algorithm requires us to (i) calculate the marginal evidence from each edge $M_{ij}(z,z')$, (ii) update messages $\mess_{j\rightarrow i}(z_i)$ between weighted edges, (iii) approximate messages $\mess_{\rightarrow i}(z_i) = \mu_i(z_i)$ between non-edges, and (iv) caclulate the vertex label probabilities $\mu_i$.

We calculate the marginal evidence $M$
\begin{equation}
\label{eq:evidswg}
\evid_{ij}(z,z') = \exp\left(T(A_{ij})\cdot \left\langle \eta \right\rangle_{R(z,z')} + T(A_{ji})\cdot \left\langle \eta \right\rangle_{R(z',z)}\right) \enspace,
\end{equation}
for each weighted edge $ij \in W$ for all $z,z'$. This takes $O(K^2 m_W)$ time. Note that $M_{ij} = M_{ji}$.
For the non-edges, we again exploit the fact that the last dimension of $T_e$ is 1 for observed edges and only need to calculate $\evid_{ij} = \evid_N$ once using Eq.~\eqref{eq:evidswg} for all non-edges $ij \in N$. 

The messages between weighted edges are
\begin{equation}
\mess_{i \rightarrow j}(z') \propto \mu_0(z') \prod\limits_{k \neq j, k \in \partial i_W } \sum\limits_{z_k} \mess_{j \rightarrow k}(z_k) \evid_{jk}(z',z_k) \enspace.
\end{equation}
Each step requires $O(|\partial i_W| K^2)$ calculations. In the case of a sparse graph, $\partial i_W=O(1)$ and since we repeat this step for each pair $i,j$ in $W$, the overall running time is $O(K^2 m_W)$.

Since there are $O(n^2)$ non-edges, the messages between non-edges must be approximated for our algorithm to be efficient. 
The idea behind this approximation is to exploit the sparsity of the weighted edges. 

To be concrete, suppose we select the Bernoulli distribution for our edge-existence distribution $f_e(x\,|\,p)$.
Then our marginal evidence takes the form   
\begin{equation}
\tilde{\evid}_{ij}(z,z') = \begin{cases}
\exp\left(\left\langle \log p \right\rangle_{z,z'}\right) \cdot \evid_{ij}(z,z')  &\text{ if } ij \in E \\
\exp\left(\left\langle \log(1-p) \right\rangle_{z,z'}\right)  &\text{ otherwise}\enspace ,
\end{cases} 
\end{equation}
where $p$ is the edge-existence parameter $\theta_e$.
If the graph is sparse, then $\left\langle\log(1-p)\right\rangle_r =O(1/n)$. 
Thus for $i,j \in E$, we have $\tilde{\evid}_{ij} \approx 1$. And therefore messages between non-edges can be approximated as
\begin{equation}
{\mess}_{i \rightarrow j}(z') = \priormu_j(z') \prod_{k \neq i} \sum_{z} \tilde{\evid}_{k,j}(z,z') {\mess}_{j \rightarrow k}(z) \approx \priormu_j(z') \prod_{k} \sum_{z} \tilde{\evid}_{k,j}(z,z') {\mess}_{j \rightarrow k}(z) = \mu_j(z') \enspace.
\end{equation} 
Thus we can approximate all messages between non-edges ${\mess}_{i \rightarrow j}(z')$ with their marginal distribution $\mu_j(z')$ taking $O(nK)$ space and time. 

The Poisson and degree-corrected case are more complicated and should follow along the lines of~\cite{yan_model_2012}. This extension is left for future work. 

Given the messages and marginal evidence, we calculate the vertex label probabilities with
\begin{align}
\mu_i(z) &\propto \priormu_i(z) \prod_{ij \in W}\sum_{z'} \evid_{ij}(z,z') \mess_{i \rightarrow j}(z') \cdot \prod_{ij \in N}\sum_{z'} \evid_N(z,z')\mu_j(z') \nonumber \\
 &= \prod_{j \in\partial i_W}\frac{\sum_{z'} \evid_{ij}(z,z') \mess_{i \rightarrow j}(z')}{\sum_{z'} \evid_N(z,z')\mu_j(z')} \cdot \left(\sum_{z'} \evid_N(z,z') \sum_j \mu_j(z')\right)^{\partial i_E} \enspace,
\end{align} 
where $\partial i_E$ is the total (in- and out-) degree of observed edges 
 and $\evid_N$ is the marginal evidence of a non-edge. Each of these updates takes $O(|\partial i_W| K^2)$ calculations.
Let $\partial i_W$ be the in and out neighborhood of $i$. 
In the case of a sparse graph, $\partial i_W$ is $O(1)$ and since we repeat this step for each pair $i,j$ in $W$, the overall running time is $O(K^2 m_W)$.

In conclusion, all three steps take $O(nK^2)$ time when the number of weighted edges and missing edges is sparse ($|W| = O(n)$ and $|M| = O(n)$). 
Although both the variational Bayes algorithm and the loopy BP algorithm have the same asymptotic running time, the constant in front of $O(nK^2)$ for the loopy BP algorithm depends on the average weighted degree of the network.


\bibliographystyle{abbrv}
\bibliography{WSBM_JCN_2014_bib}

%


%
%

\end{document}